\documentclass[runningheads]{llncs}

\usepackage{eccv}              

\usepackage{eccvabbrv}

\usepackage{url}
\usepackage{booktabs}
\usepackage{amsfonts}
\usepackage{nicefrac}
\usepackage{microtype}
\usepackage{multirow}
\usepackage{graphicx}
\usepackage{csvsimple}
\usepackage{tabularx}
\usepackage{siunitx}
\usepackage{makecell}
\usepackage{pifont}
\usepackage{threeparttable}
\usepackage{svg}
\usepackage{colortbl}
\usepackage{listings}
\definecolor{heatmapColor}{HTML}{63A4FF}

\usepackage{array}
\usepackage{verbatim}

\definecolor{lightgray}{HTML}{777777}
\definecolor{checkcolor}{HTML}{0000FF}

\newcommand{\bst}[1]{%
  \cellcolor{green!15}
  {\emph{#1}}%
}

\newcommand{\sbst}[1]{%
  \cellcolor{orange!15}
  {#1}%
}

\newif\ifoodrow
\newcommand{\oodrow}{\global\oodrowtrue}
\newcolumntype{L}{>{\ifoodrow\cellcolor{blue!6}\fi}l}
\newcolumntype{C}{>{\ifoodrow\cellcolor{blue!6}\fi}c}
\newcolumntype{E}{>{\ifoodrow\cellcolor{blue!6}\fi}c<{\global\oodrowfalse}}

\makeatletter
\renewcommand\paragraph{\@startsection{paragraph}{4}{\z@}%
                       {-12\p@ \@plus -4\p@ \@minus -4\p@}%
                       {-0.5em \@plus -0.22em \@minus -0.1em}%
                       {\normalfont\normalsize\bfseries}}
\makeatother

\usepackage[accsupp]{axessibility}

\usepackage{hyperref}

\usepackage{orcidlink}

\title{OmniFall: From Staged Through Synthetic to Wild, A Unified Multi-Domain Dataset for Robust Fall Detection}

\titlerunning{OmniFall: A Unified Multi-Domain Dataset for Robust Fall Detection}

\author{David Schneider\inst{1}\thanks{Corresponding author: david.schneider@kit.edu} \and
  Zdravko Marinov\inst{1} \and
  Moritz Mistol\inst{1} \and
  Zeyun Zhong\inst{1} \and
  Alexander Jaus\inst{1} \and
  Rodi D\"uger\inst{1} \and
  Rafael Baur\inst{1} \and
  M. Saquib Sarfraz\inst{2} \and
Rainer Stiefelhagen\inst{1}}

\authorrunning{D.~Schneider et al.}

\institute{Karlsruhe Institute of Technology \and
Mercedes-Benz Tech Innovation}

\begin{document}

\maketitle

\begin{abstract}
  Visual fall detection models are usually trained on small, staged datasets.
  Their real-world utility remains unclear; such data lacks
  diversity and evaluation protocols differ from paper to paper.
  We propose OmniFall, a unified benchmark of 15k videos
  (80\,hours) with frame-level annotations in a single 16-class
  taxonomy. It spans three domains: OF-Staged unifies eight staged
  datasets with cross-subject and cross-view splits; OF-Synthetic
  adds 12k videos (17\,h) with controlled demographic and environmental
  diversity; and OF-In-the-Wild provides a test-only set of genuine
  accident videos. We evaluate fine-tuned models as well as much larger
  zero-shot multimodal LLMs. On in-the-wild fall events, both do
  comparably well. The clinically critical \emph{fallen} state is where
  they part: zero-shot models keep confusing \emph{fallen} with
  \emph{lying}, whereas models fine-tuned on synthetic data
  with explicit fallen-state scenes do substantially better.
  We release the unified annotations, the
  synthetic data, and the in-the-wild test set to foster the development of fall and
  fallen-state detectors for uncontrolled environments.\\
  Dataset:  \url{https://hf.co/datasets/simplexsigil2/omnifall}
\end{abstract}

\section{Introduction}\label{sec:intro}

Falls remain a leading cause of fatal and non-fatal injury in older
adults, claiming about 684,000 lives globally and sending more than
37 million people to medical care, yearly~\cite{WHO2021Falls}.
Preventing death hinges on detecting such falls, but even more so on
recognizing the sustained \emph{fallen state}, when a person remains
on the ground unable to summon help. "Long-lie" episodes increase
one-year mortality
significantly~\cite{tinetti1993predictors,fleming2008inability}.
Beyond elderly care, falls pose risks in workplaces,
public spaces, and during recreational activities, often under
diverse, uncontrolled conditions that static indoor datasets cannot
represent.

Computer vision-based fall detection could automate monitoring in
care environments, yet faces critical limitations: (1) systems target
brief falls rather than persistent fallen states; (2) models trained
on small, homogeneous
datasets~\cite{ur-fall,cauca,edf-occu,gmdcsa,le2i,mcfd,cmdfall,up-fall}
generalize poorly across environments; (3) collecting fall data from
vulnerable elderly populations raises ethical and practical barriers;
(4) no standardized benchmark evaluates genuine accidents or
cross-domain generalization; and (5) Multimodal Large Language Models (MLLMs) are powerful
generalist vision models, yet underexplored for fall detection.

\begin{figure}[!t]
  \centering
  \includegraphics[width=0.85\linewidth]{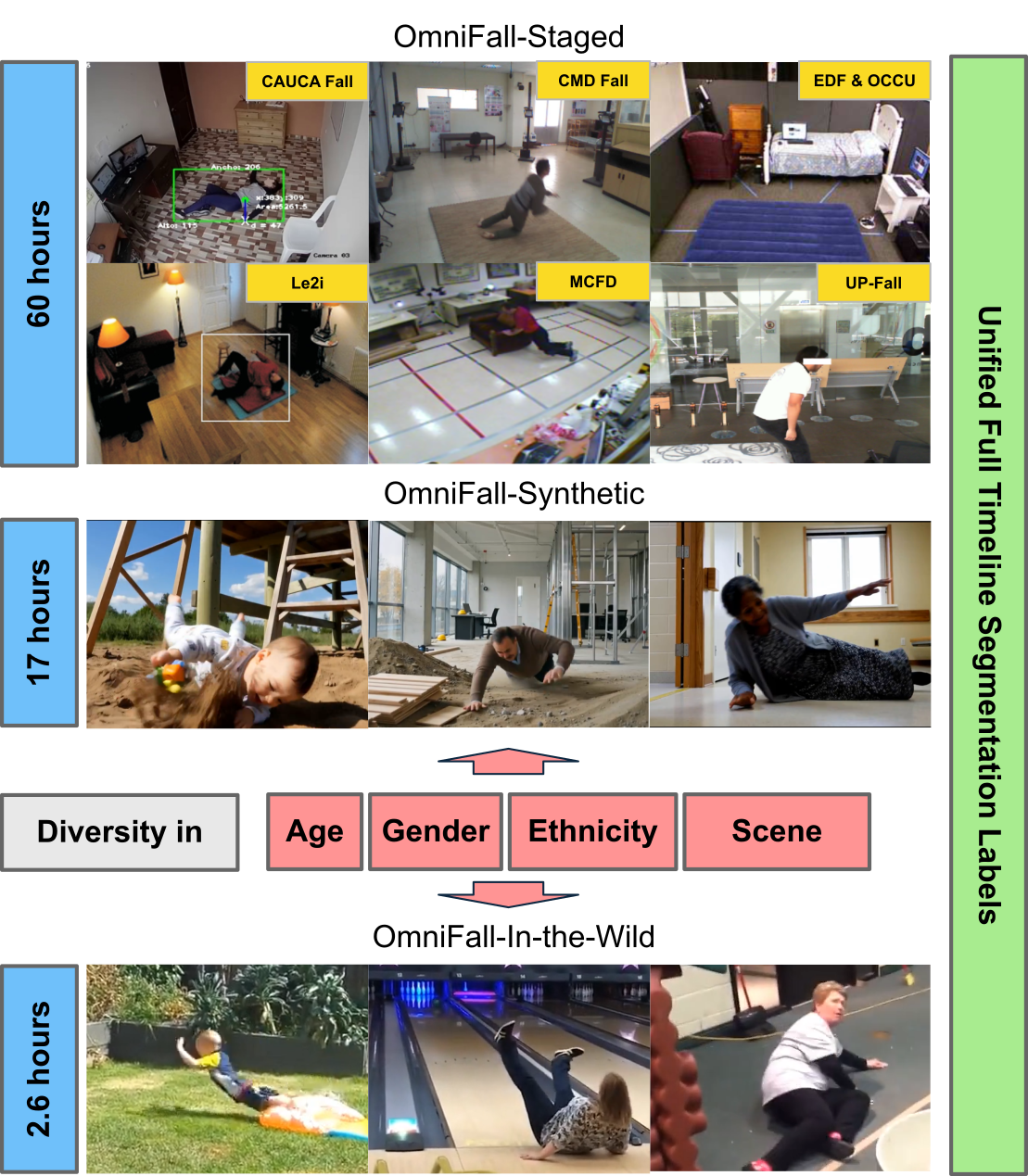}
  \caption{OmniFall comprises OF-Staged, OF-Synthetic and
  OF-In-the-Wild with cross-dataset compatible labels.}
  \label{fig:examples}
\end{figure}

We introduce \emph{OmniFall}, a unified benchmark comprising three
complementary components: \emph{OmniFall-Staged} (\emph{OF-Sta}: staged
real-world data), \emph{OmniFall-In-the-Wild} (\emph{OF-ItW}: genuine
accident videos), and \emph{OmniFall-Synthetic} (\emph{OF-Syn}: large-scale
synthetic data). The benchmark provides 33 hours of densely annotated footage.
Five of the eight staged datasets include synchronized multi-view
cameras; counting all camera feeds raises the total to 82 hours.
This scale enables systematic cross-domain evaluation previously
impossible due to annotation heterogeneity. A clinically motivated
16-class taxonomy (\cref{sec:dataset}) supports detection of critical
fallen states even when the fall itself is missed.

\textbf{Why unify existing staged datasets?} Staged datasets cover
few environments and subjects with poor generalization to new domains. \emph{OmniFall-Staged} merges eight such
datasets~\cite{cmdfall,up-fall,le2i,gmdcsa,edf-occu,mcfd,cauca} spanning 101 subjects and 29 camera views under
one annotation scheme. Videos are
frame-wise re-annotated allowing for cross-subject and cross-view evaluations across dataset boundaries.

\textbf{Why the need for in-the-wild evaluation?} Staged datasets display healthy actors performing falls in controlled settings with safety equipment, so their generalization to real-world deployment remains unknown. \emph{OmniFall-In-the-Wild (OF-ItW)} addresses this with genuine
accident videos from uncontrolled environments, held out as a wild
test target whose 1.8\,h test portion alone exceeds the total duration
of many existing fall detection datasets.

\textbf{Why add synthetic data?} Recording real falls of diverse demographic populations including the elderly and children raises ethical concerns and practical barriers.
\emph{OmniFall-Synthetic (OF-Syn)} sidesteps both with generated
video offering controlled demographic and environmental variation at
no privacy cost. Critically, it explicitly models extended post-fall
ground contact, capturing the clinically important \emph{fallen} state
that staged recordings underrepresent, since actors there stand up
immediately.

Our contributions are threefold.
\textbf{(1)} We manually re-annotate eight public fall detection
datasets frame by frame in a clinically
motivated 16-class taxonomy, and define the first standardized
staged-to-wild protocol with cross-subject/view
splits and a held-out in-the-wild test set of genuine accidents.
\textbf{(2)} OF-Synthetic
adds synthetic-to-wild evaluation and
diverse training data with controlled demographic variation.
\textbf{(3)} We test both fine-tuned models
and zero-shot  MLLMs and find they fail differently:
both reach similar fall-event detection on genuine
accidents, but zero-shot MLLMs systematically miss the clinically
critical \emph{fallen} state. Fine-tuned models trained on synthetic
data, which shows extended post-fall contact, detect the
fallen state best across domains. The distinction thus appears to need
task-specific training, rather than model scale alone.

\section{Related Work}\label{sec:related}

Many fall detection datasets have been introduced in recent years
\cite{cmdfall,up-fall,le2i,gmdcsa,edf-occu,mcfd,cauca,fallfree,ur-fall,high-quality-fd,act42,simple},
reflecting the growing interest in automated fall detection.

\begin{table}[t]
  \centering
  \setlength{\tabcolsep}{1.5pt}
  \small
  \caption{Datasets in the OmniFall benchmark. The first eight rows
    constitute OF-Staged (OF-Sta), followed by OF-In-the-Wild (OF-ItW)
  and OF-Synthetic (OF-Syn).}
  \label{tab:fall_detection_datasets}

  \begin{threeparttable}
    \begin{tabular}{@{}lcrccrcrrrrr@{}}
      \toprule
      \multirow{2}{*}{\textbf{Dataset}} &
      \multirow{2}{*}{\textbf{Year}} &
      \multirow{2}{*}{\textbf{Vids}} &
      \multirow{2}{*}{\textbf{Res.}} &
      \multirow{2}{*}{\textbf{FPS}} &
      \multirow{2}{*}{\textbf{Sub}} &
      \multirow{2}{*}{\makecell{\textbf{Views}\\\textbf{(Sync.)}}} &
      \multirow{2}{*}{$\bar{t}$\,\textbf{(s)}} &
      \multicolumn{2}{c}{\textbf{Segments}} &
      \multicolumn{2}{c}{\textbf{Dur.\,(h)}} \\
      \cmidrule(lr){9-10} \cmidrule(lr){11-12}
      & & & & & & & &
      \multicolumn{1}{c}{\textbf{SV}} &
      \multicolumn{1}{c}{\textbf{MV}} &
      \multicolumn{1}{c}{\textbf{SV}} &
      \multicolumn{1}{c}{\textbf{MV}} \\
      \midrule
      MCFD~\cite{mcfd}        & 2010           & 192   & 480p &
      $\leq$30 & 1  & 8\,(\checkmark)  & 4.3  & 169    & 1,352  & 0.2  & 1.6  \\
      Le2i~\cite{le2i}        & 2013           & 190   & 240p & 25
      & 9  & 6\,(\ding{55})   & 3.0  & 967    &        & 0.8  &      \\
      EDF~\cite{edf-occu}     & 2014           & 10    & 240p & 30
      & 5  & 2\,(\checkmark)  & 3.1  & 254    & 508    & 0.2  & 0.4  \\
      OCCU~\cite{edf-occu}    & 2014           & 10    & 240p & 30
      & 5  & 2\,(\ding{55})   & 3.5  & 245    & 484    & 0.3  & 0.5  \\
      CMDFall~\cite{cmdfall}  & 2018           & 384   & 480p & 20
      & 50 & 7\,(\checkmark)  & 4.3  & 6,026  & 42,143 & 7.1  & 49.8 \\
      UP Fall~\cite{up-fall}  & 2019           & 1,118 & 480p & 18
      & 17 & 2\,(\checkmark)  & 13.6 & 1,213  & 2,426  & 4.6  & 9.2  \\
      CAUCA~\cite{cauca}      & 2022           & 100   & 480p & 23
      & 10 & 1                & 3.9  & 258    &        & 0.3  &      \\
      GMDCSA~\cite{gmdcsa}    & 2024           & 160   & 720p & 30
      & 4  & 3\,(\ding{55})   & 2.8  & 458    &        & 0.4  &      \\
      \midrule
      \textbf{OF-ItW}         & 2026$^\dagger$ & 818   & var. & 30
      &    & {>}818           & 2.4  & 4,022  &        & 2.7  &      \\
      \textbf{OF-Syn}         & 2026           & 12k   & 720p & 16
      &    & 12k              & 3.2  & 19,228 &        & 16.9 &      \\
      \bottomrule
    \end{tabular}
    \begin{tablenotes}[flushleft]
      \footnotesize
    \item $\dagger$ With videos from~\cite{oops}, 2019. \textbf{Sub:} subjects,
      \textbf{Sync.:} synchronized views,
      \textbf{SV/MV:} single-view\,/\,multi-view,
      $\bar{t}$: mean segment duration.
    \end{tablenotes}

  \end{threeparttable}

\end{table}

\begin{figure}[tbp]
  \centering
  \includegraphics[width=0.95\linewidth]{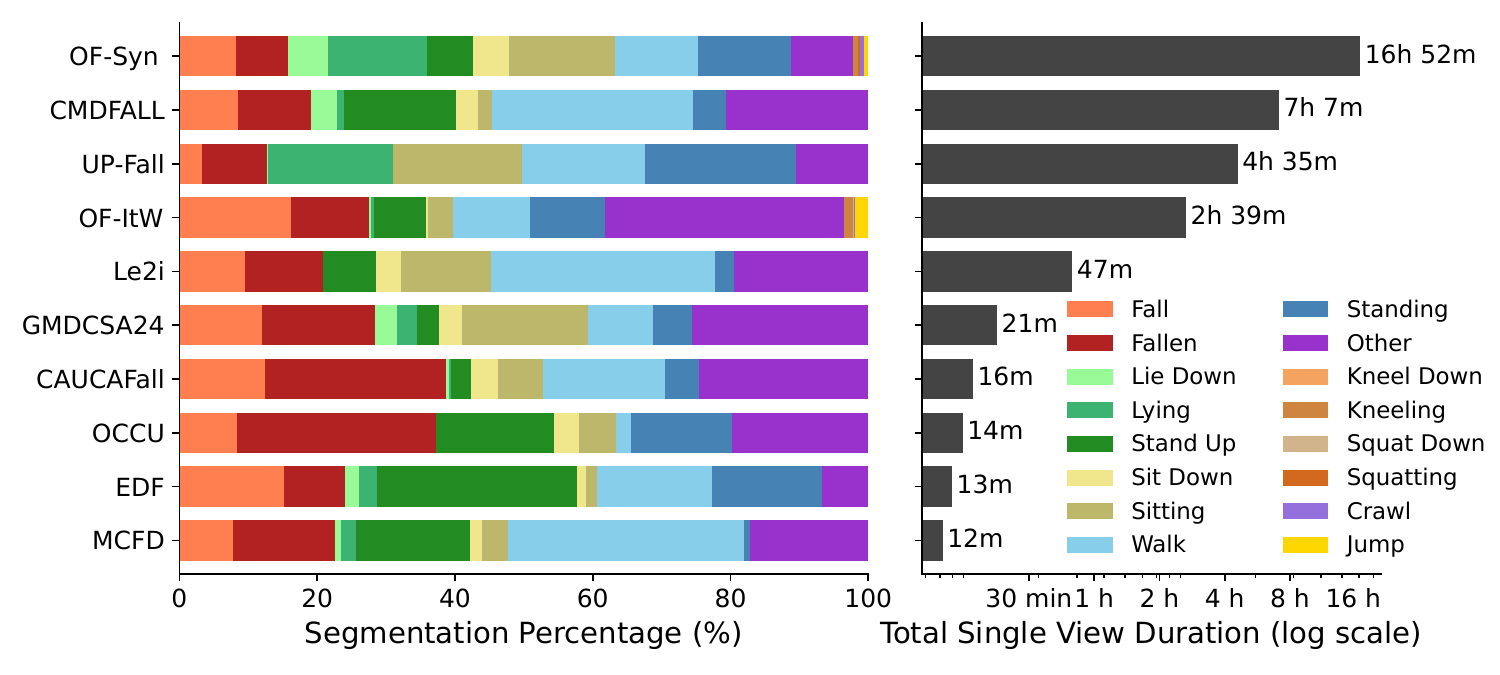}
  \caption{Label distribution and single-view duration per dataset:
  eight OF-Sta constituents, OF-ItW, and OF-Syn (33h total).}
  \label{fig:dataset-sizes}
\end{figure}

\paragraph{Fall Detection Datasets in OmniFall-Staged.} To collect
existing public fall detection datasets from prior work, we conducted
a systematic search in accordance with PRISMA guidelines
(5)-(9)~\cite{page2021prisma}. Of the 98 surveys retrieved, only 8
met all inclusion criteria. These datasets (1)-(8) are listed in Table~\ref{tab:fall_detection_datasets}.

(1) The \textbf{CMDFall} dataset~\cite{cmdfall} is a multimodal,
multiview collection acquired from seven synchronized Kinect sensors
(providing color and depth data) together with two accelerometers. It
comprises recordings of 50 subjects performing eight distinct fall
types alongside twelve non-fall activities.
(2) \textbf{UP Fall}~\cite{up-fall} offers a multimodal setup from 17
subjects in a laboratory environment. It includes vision data from a
pair of synchronized cameras (640×480 at 18 fps), inertial
measurements from accelerometers and gyroscopes worn at different body locations
covering six activities of daily living (ADL) and five fall types.
(3) The \textbf{Le2i} dataset~\cite{le2i} contains 143 staged falls
and 79 ADL actions recorded from a single fixed camera at 640×480
pixels and 25 fps across four environments and six views.
(4) \textbf{GMDCSA}~\cite{gmdcsa} delivers high temporal resolution
data, capturing 81 fall and 79 ADL clips recorded with a stationary
laptop camera at 1280×720 pixels and 30fps in three distinct home
settings. Its design emphasizes intra-class variability, with
subjects altering clothing and backgrounds between clips. (5) The
\textbf{EDF} and (6) \textbf{OCCU} datasets~\cite{edf-occu} were
acquired in a controlled apartment environment using two Kinect
sensors at 320×240 pixels and 30 fps. The EDF set focuses on 40
non-occluded falls by five subjects performing falls from eight
directions, as well as 30 non-fall actions. In comparison, OCCU
presents a more challenging scenario in which 30 falls become
partially occluded near the end, accompanied by a suite of 80 ADL
actions. (7)\textbf{ The Multiple Camera Fall (MCFD)}
dataset~\cite{mcfd,auvinet2011fall} leverages eight synchronized IP cameras installed
in a fixed indoor setting to capture 22 fall scenarios,
facilitating 3D reconstructions either explicitly or implicitly. (8)
The \textbf{CAUCA Fall} dataset~\cite{cauca} provides recordings from
10 subjects in an apartment setting under varying lighting
conditions. Captured at 720×480 resolution and 23 fps, it includes
both color and infrared modes (in low-light situations) with
detailed, frame-wise annotations that report not only fall versus
non-fall labels but also spatial metrics, such as distance and angle
with respect to the camera. \textbf{OOPS!}~\cite{oops} is a 20.7
k-clip (>50 h) video dataset with labels marking when intentional
actions turn accidental, supporting benchmarks in intention
classification and accident forecasting.

\paragraph{Other Datasets.} We identified several additional fall
detection datasets, but excluded them from OF-Staged due to specific
limitations. UR Fall~\cite{ur-fall} was not included due to its
limited size. MUVIM~\cite{muvim} offers multimodal data, but all
videos are captured from a top-down view. FPDS~\cite{fpds} consists
of static images, we only considered full video datasets. While High
Quality FSD~\cite{high-quality-fd}, FallFree~\cite{fallfree},
SIMPLE~\cite{simple}, and ACT4$^2$~\cite{act42} align well with our
objectives, we were unable to include them due to missing usage
licenses or the lack of usage permission.

\paragraph{What Is Missing.}
These datasets comprise \emph{staged} recordings of healthy
participants in controlled environments. They are useful for method
development, but none shows how models transfer to
\emph{in-the-wild} falls, and their heterogeneous
label sets and protocols make cross-dataset comparison hard. Three
gaps stand out:
(1) \emph{Staging bias:} lab scenes, fixed cameras, and scripted
routines dominate.
(2) \emph{Annotation heterogeneity:} labels range from binary
fall/no-fall to coarse ADL categories. Few offer dense, consistent
timelines. To reconcile these schemes, every frame
must be re-annotated under a domain-appropriate
taxonomy, which no prior work has done at scale.
(3) \emph{Lack of standardized OOD evaluation:} no widely-used
protocol quantifies staged$\rightarrow$wild generalization.

\textbf{OmniFall}
(Table~\ref{tab:fall_detection_datasets}) addresses all three: we
unify eight staged datasets by re-annotating them framewise (OF-Staged), curate an in-the-wild test
set from genuine accidents (OF-ItW), and add a diverse synthetic dataset (OF-Syn). The benchmark and its three components
are detailed in the following section.

\section{Dataset}
\label{sec:dataset}

\begin{figure}[tbhp]
  \centering
  \includegraphics[width=1\linewidth]{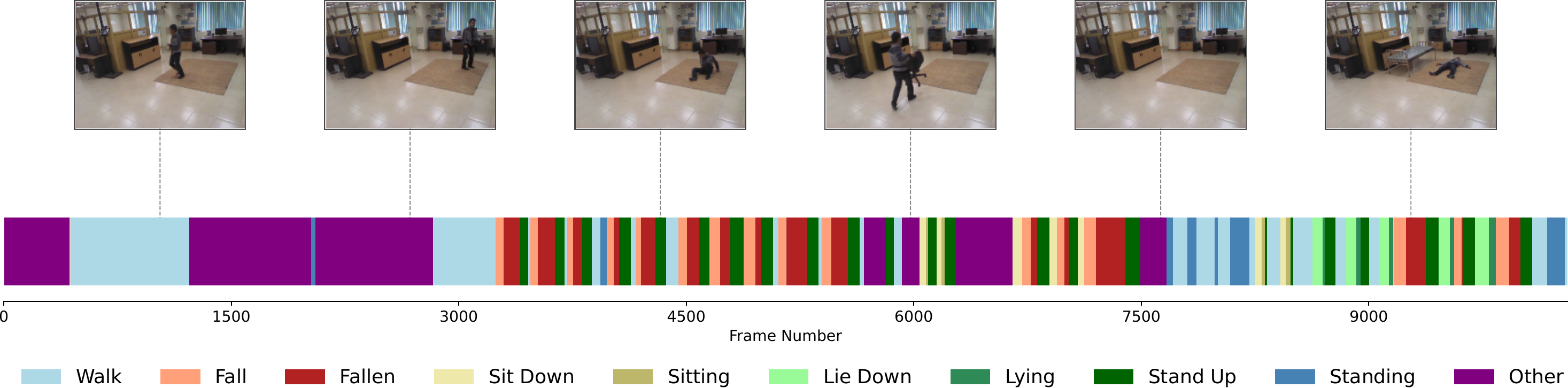}
  \caption{Our annotations on an example video of CMDFall~\cite{cmdfall}.}
  \label{fig:segmentation-example}
\end{figure}

\paragraph{Overview.}
\textbf{OmniFall} combines three complementary components, each
detailed below (\cref{tab:fall_detection_datasets},
\cref{fig:examples}): \textbf{OF-Staged} (eight public staged
datasets), \textbf{OF-In-the-Wild} (\textbf{OF-ItW}), and
\textbf{OF-Synthetic} (\textbf{OF-Syn}). Together they total
$\sim$33\,h of single-view footage and enable \emph{in-distribution}
training on staged data and \emph{out-of-distribution} (OOD) testing
on wild and synthetic.

\paragraph{Design Principles.}
Our goal is \emph{comparability across domains}. To that end we:
(i) \emph{Standardize labels} by manually (re-)annotating all datasets.
(ii) \emph{Densely annotate} full videos
(Figure~\ref{fig:segmentation-example}), so methods can be
evaluated on both classification and timeline segmentation.
(iii) \emph{Unify splits} by providing cross-subject (CS) and
cross-view (CV) partitions for each staged dataset, mirroring
original protocols when available.
(iv) \emph{Enable OOD assessment} with a held-out OF-ItW split.

\paragraph{Annotations.}
Because source datasets make varying assumptions about
fall semantics and their original annotations often cover only
isolated events rather than full video timelines, we annotate all
videos from scratch rather than mapping existing labels, avoiding
bias from preexisting schemes and ensuring complete temporal
coverage.
Our annotations are frame-wise and comprise dynamic
actions: \emph{fall, sit\_down, lie\_down, stand\_up, walk, kneel\_down, squat\_down,
crawl, jump}; six static states: \emph{fallen, sitting, lying,
standing, kneeling, squatting}; plus \emph{other} as a catch-all category for remaining frames.
Interviews with medical professionals
identified a \emph{fall}/\emph{fallen} distinction as clinically
essential for catching ``long-lie'' episodes, so our annotation taxonomy models the
\emph{fallen} state separately from the fall itself. We further include visually similar actions such as
\emph{lie\_down} and \emph{lying} since avoiding confusion between lying down and falling has great practical relevance.
Additional classes are determined by reviewing the datasets and adding major activities.
The annotation process is described in detail in the appendix.

\paragraph{OF-Staged.}
OF-Staged spans 14\,h of recordings across eight datasets
(CMDFall~\cite{cmdfall}, UP-Fall~\cite{up-fall}, Le2i~\cite{le2i}, GMDCSA24~\cite{gmdcsa},
CAUCA Fall~\cite{cauca}, EDF~\cite{edf-occu}, Occu~\cite{edf-occu} and MCFD~\cite{mcfd}, ~\cref{fig:dataset-sizes}),
with synchronized views excluded from
duration accounting but preserved for CV evaluation.
When multi-view recordings are included, CS training
includes up to 61.5\,h of footage, providing OF-Sta with a
training volume advantage over OF-Syn (16.9\,h single-view) under cross-subject
evaluation. We provide CS/CV splits per dataset and
\emph{combined} splits (OF-Staged) that support training and
testing across domains.

\paragraph{OF-In-the-Wild.}
Starting from OOPS~\cite{oops}, we filter clips containing falls, then
\emph{manually verify and densely annotate} 818 videos (2.65\,h).
These contain 1.3k segments in 10/20/70 train/val/test splits with real-world camera motion, occlusion,
and clutter. We selected clips with a clearly visible fall, and
covered indoor and outdoor scenes under varied lighting. Collecting genuine falls from elderly people is ethically
impractical; OF-ItW, with its unscripted accident videos, is a close available proxy.
In this work we almost exlusively use OF-ItW as \emph{test-only}, measuring staged$\rightarrow$wild generalization.

\paragraph{OF-Synthetic.}
OF-Syn contributes 12k videos (17\,h) generated with Wan
2.2~\cite{wan2025} at 1280$\times$720 and 16\,fps, each a five-second
clip of a unique individual and scene. Domain-gap analyses
(\S\ref{sec:eval-domain}) show OF-Syn sits closer to OF-ItW in
feature space than individual staged datasets.
This enables its use for sim$\rightarrow$real studies.

\begin{figure}[t]
  \centering
  \includegraphics[width=\linewidth]{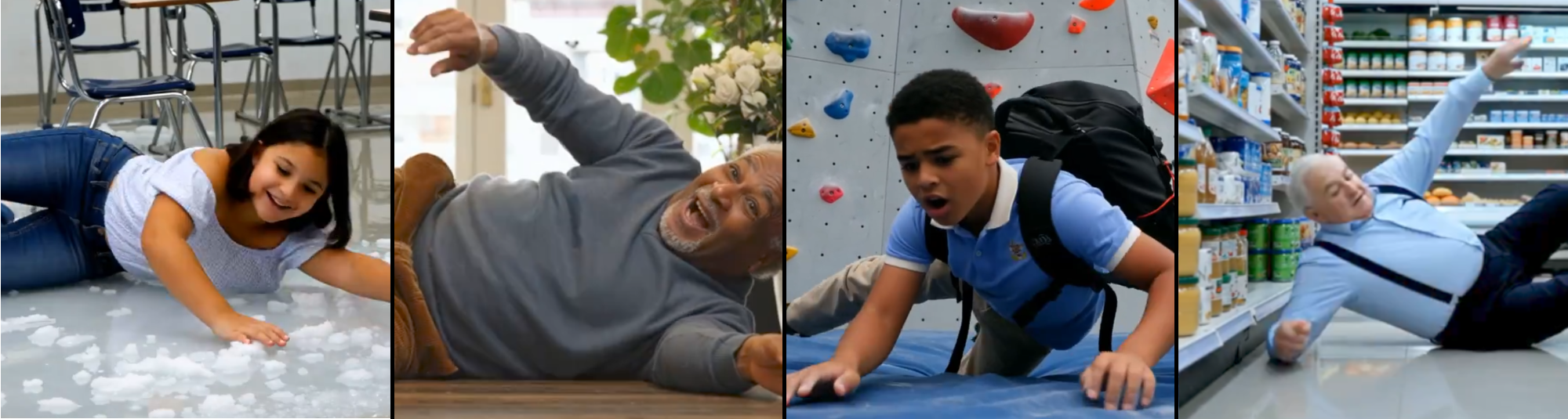}
  \caption{Expression conditioning mitigates Wan\,2.2's smile bias on
  falls: without (left) and with (right) explicit expression prompts.}
  \label{fig:smile}
\end{figure}

We build each clip from a structured prompt with varying axes:
who the person is, where they are and what they wear,
how the fall unfolds, and where the camera sits. Demographics span six
age groups from toddler to elderly and the seven OMB SPD-15 race and
ethnicity categories~\cite{OMB2024SPD15}, alongside body type and
gender, while scenes draw on over a thousand distinct environments.
Each fall combines one of 30 mechanisms (e.g., a slip on ice) with a
direction and a recovery attempt, and the camera ranges from eye-level
to CCTV and top-down views.
We fixed specific Wan\,2.2 failure modes with prompt engineering, such as
a real-time frame-rate cue reducing slow-motion artifacts, and
conditioning on expressions such as \emph{startled} or \emph{panicked}
suppressing smiling faces during falls (Fig.~\ref{fig:smile}). We screened all 12{,}000 clips
and re-ran generation for clips with unsafe content, implausible motion, or
artifacts. Additional details in the appendix.

\section{Experiments}\label{sec:exp}

Our evaluation comprises three complementary experiments using
different methodologies. For domain shift analysis
(\cref{sec:eval-domain}), we extract frozen motion oriented features
from I3D~\cite{carreira2017quo} and
VideoMAE-K400~\cite{tong2022videomae} and more appearance-based
features from SigLIP2~\cite{tschannen2025siglip} to analyze dataset
similarities and feature clustering. For action and fall
classification (\cref{sec:action_classification}), we fine-tune
VideoMAE-K400 end-to-end on our benchmark datasets. Timeline
segmentation follows established protocols using SigLIP2
features (\cref{sec:time-seg}). Additional details
in the appendix.
\begin{figure*}[t]
  \centering
  \includegraphics[width=\linewidth]{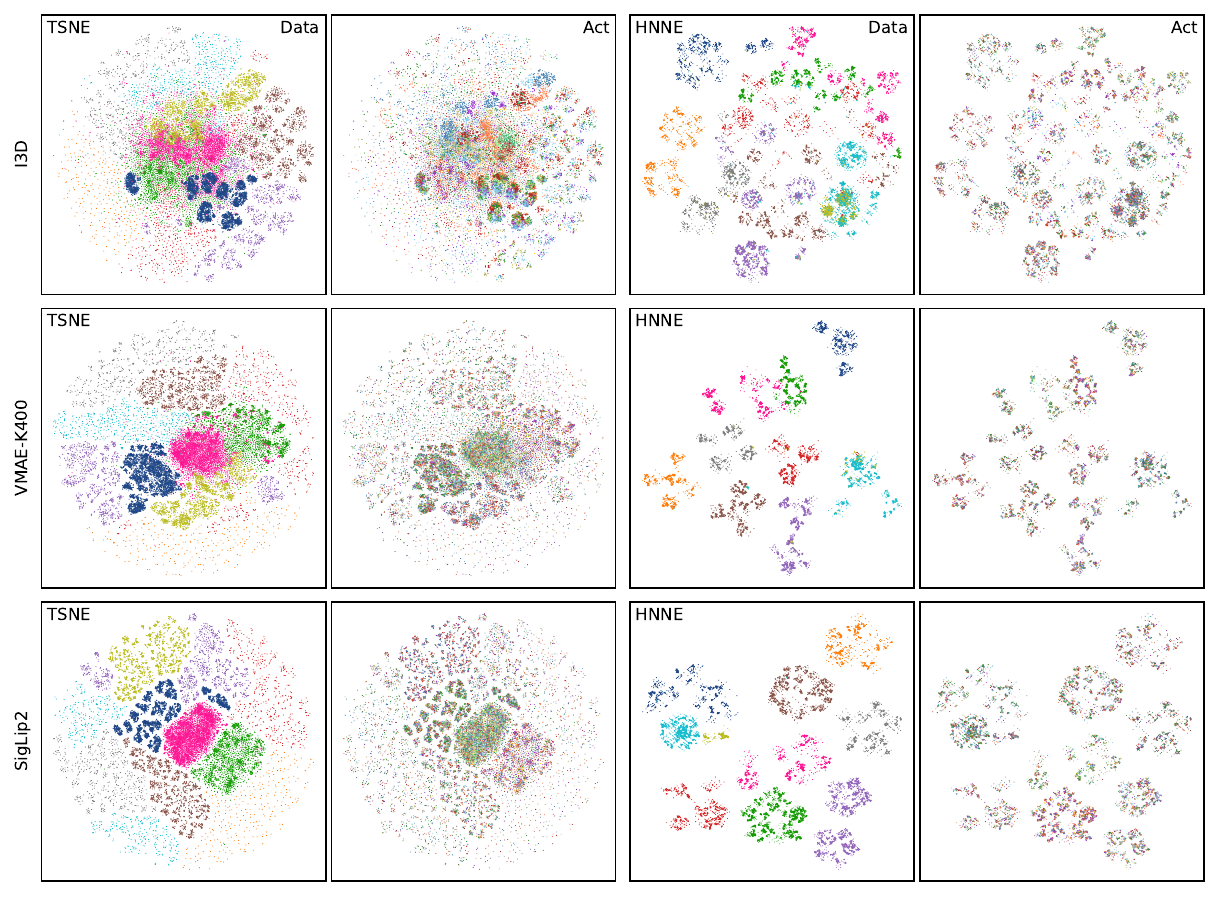}\\
  \includegraphics[width=\linewidth]{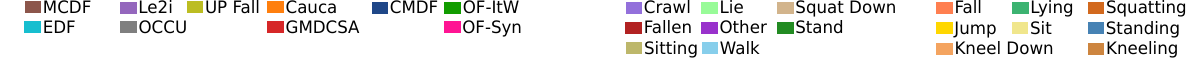}
  \caption{2D projections of frozen embeddings (I3D, VideoMAE-K400,
    SigLIP2), color-coded by dataset (odd columns) and activity (even
    columns). Rows correspond to feature type; columns show t-SNE and
  h-NNE clustering. Full-size plots in the appendix.}
  \label{fig:tsne-hnne}
\end{figure*}

\begin{figure*}[t]
  \centering
  \includegraphics[width=\linewidth]{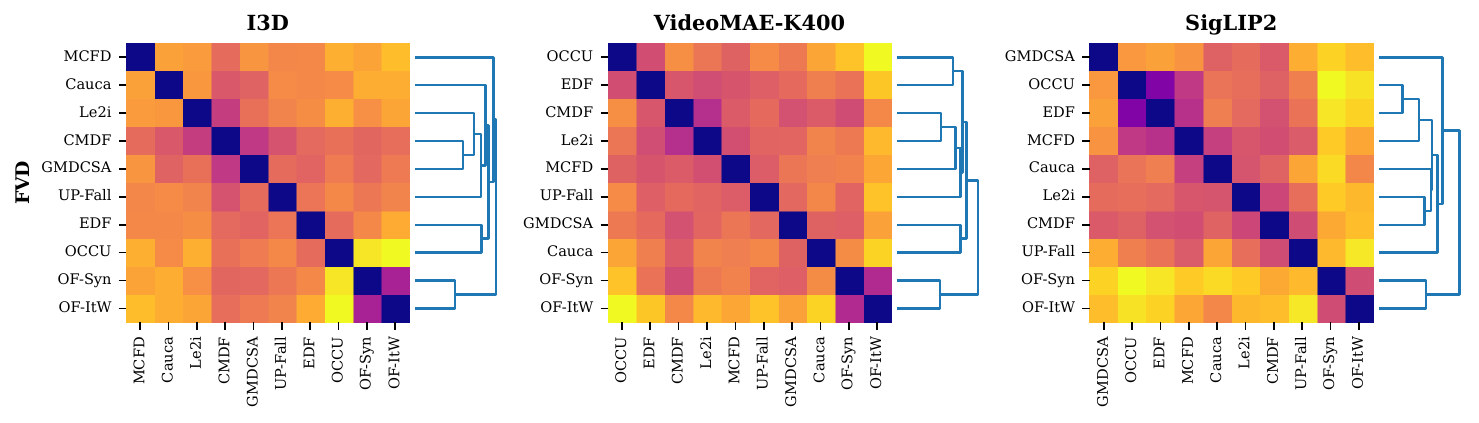}
  \caption{Pairwise Fréchet Video Distances calculated on features
    from I3D, VideoMAE-K400 and SigLIP2, visualised as heat‑maps.
    Darker colors indicate higher similarity (smaller distances)
  between datasets. Numerical values are listed in the appendix.}
  \label{fig:domain-heatmaps}
\end{figure*}

\subsection{Domain Shift Analysis}\label{sec:eval-domain}
\paragraph{Feature-Space Visualization.}
Figure~\ref{fig:tsne-hnne} visualizes frozen embeddings from
I3D, VideoMAE-K400, and SigLIP2 via t-SNE and h-NNE projections
(5{,}000 samples per domain, five-frame temporal means). Across all
backbones, domains form largely disjoint clusters, confirming a
strong dataset shift between staged, in-the-wild, and synthetic
videos. The separation is most pronounced for SigLIP2 and clearly
visible for VideoMAE and I3D as well.

Figure~\ref{fig:domain-heatmaps} shows pairwise Fr\'{e}chet Video
Distances across three feature types.
CMDFall demonstrates high similarity (darker colors) to most other
staged datasets, consistent with its broad coverage of fall types and
ADL categories. OF-Syn and OF-ItW exhibit strong mutual similarity,
likely because both contain a large proportion of outdoor activities
absent from any staged dataset. This domain overlap partly explains
OF-Syn's transfer advantage in
\cref{sec:action_classification}. It also illustrates a practical
strength of synthetic generation: diverse environments can be covered
at negligible cost, whereas staged recordings are typically confined
to laboratory settings. Conversely, OF-Staged shows substantial
dissimilarity to OF-ItW on VMAE-K400, corresponding to the OOD
performance gaps in Section~\ref{sec:action_classification}.
We encourage cross-referencing these distance-based visualizations
with the dataset examples provided in the supplementary material.

\subsection{Action and Fall Classification}
\label{sec:action_classification}

\begin{figure*}[t]
  \centering
  \includegraphics[width=\linewidth]{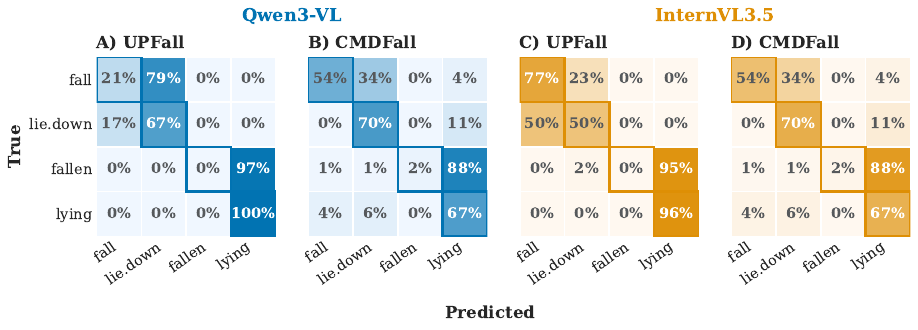}
  \caption{Zero-shot confusion matrices on UP-Fall and
    CMDFall for Qwen3-VL-8B and InternVL3.5-8B, restricted to the
    semantically adjacent classes \emph{fall}, \emph{lie down},
    \emph{fallen}, and \emph{lying}. Both models confuse \emph{fallen} with \emph{lying} and, to varying degrees,
  \emph{fall} with \emph{lie down}.}
  \label{fig:conf-matrices}
\end{figure*}

\begin{table}[t]
  \centering
  \caption{Per-dataset \textbf{in-domain} (ID) vs.\ \textbf{in-the-wild}
    (ItW) generalization. A VideoMAE-K400 model is trained on each
    staged dataset and tested on its own test split (ID) and on OF-ItW.
  \textbf{Se}nsitivity, \textbf{Sp}ecificity, \textbf{F1} score.}
  \label{tab:per-dataset-itw}
  \renewcommand{\arraystretch}{0.9}
  \setlength{\tabcolsep}{4pt}
  \small
  \begin{threeparttable}
    \begin{tabular}{ll rrr c rrr}
      \toprule
      \multirow{2}{*}{Test} & \multirow{2}{*}{Train dataset} &
      \multicolumn{3}{c}{Fall} & & \multicolumn{3}{c}{Fallen} \\
      \cmidrule(lr){3-5}\cmidrule(lr){7-9}
      & & Se & Sp & F1 & & Se & Sp & F1 \\
      \midrule
      \multirow{7}{*}{\rotatebox{90}{\textbf{ID}}}
      & CMDFall   & 91.2 & 99.3 & 93.0 && 92.0 & 98.7 & 90.9 \\
      & Up-Fall   & 57.9 & 99.8 & 72.9 && 40.4 & 99.8 & 57.1 \\
      & Le2i      & 100  & 100  & 100  && 100  & 99.5 & 97.7 \\
      & GMDCSA24  & 76.5 & 93.4 & 74.3 && 76.5 & 100  & 86.7 \\
      & EDF       & 100  & 100  & 100  && 65.0 & 100  & 78.8 \\
      & OCCU      & 100  & 97.7 & 94.1 && 100  & 100  & 100  \\
      & CaucaFall & 100  & 94.7 & 90.0 && 100  & 100  & 100  \\
      \midrule
      \multirow{7}{*}{\rotatebox{90}{\textbf{OF-ItW}}}
      & CMDFall   & 51.1 & 86.9 & 52.3 && 10.0 & 98.7 & 16.6 \\
      & Up-Fall   & 9.5  & 99.6 & 16.9 && 1.6  & 99.8 & 3.2  \\
      & Le2i      & 56.8 & 94.8 & 58.8 && 50.5 & 86.7 & 33.1 \\
      & GMDCSA24  & 37.8 & 96.7 & 47.0 && 12.3 & 97.7 & 17.7 \\
      & EDF       & 15.6 & 98.6 & 24.8 && 18.9 & 94.8 & 21.0 \\
      & OCCU      & 8.7  & 98.4 & 14.5 && 1.5  & 98.9 & 2.7  \\
      & CaucaFall & 8.0  & 98.9 & 13.8 && 11.2 & 97.7 & 16.2 \\
      \bottomrule
    \end{tabular}
    \begin{tablenotes}
      \scriptsize
    \item[\ ] Each row reports one VideoMAE-K400~\cite{tong2022videomae}
      model trained on that dataset's training split. ID: same
      dataset's held-out test split; ItW: the OF-ItW test set.
    \end{tablenotes}
  \end{threeparttable}
\end{table}

\begin{table}[!tbp]
  \centering
  \renewcommand{\arraystretch}{0.88}
  \caption{Evaluation of zero-shot MLLMs and differently fine-tuned VMAE-K400 models on OF-Staged (CS test splits).
    \textbf{B}al.\ \textbf{Acc}uracy, \textbf{Se}nsitivity,
  \textbf{Sp}ecificity.}
  \label{tab:unified-classification}
  \resizebox{\columnwidth}{!}{%
    \begin{threeparttable}
      \setlength{\tabcolsep}{1.5pt}
      \begin{tabular}{l L L | CCC C CCC C CCC C CCE}
        \toprule
        \multirow{2}{*}{Test} &
        \multirow{2}{*}{Model} &
        \multirow{2}{*}{Train} &
        \multicolumn{3}{c}{16-class} & &
        \multicolumn{3}{c}{Fall $\Delta$} & &
        \multicolumn{3}{c}{Fallen $\Delta$} & &
        \multicolumn{3}{c}{Fall $\cup$ Fallen $\Delta$} \\
        \cmidrule(lr){4-6}\cmidrule(lr){8-10}\cmidrule(lr){12-14}\cmidrule(lr){16-18}
        & & & BAcc & Acc & F1 & &
        Se & Sp & F1 & &
        Se & Sp & F1 & &
        Se & Sp & F1 \\
        \midrule
        \oodrow
        \multirow{6}{*}{\rotatebox{90}{CMDF~\cite{cmdfall}}}
        & Qwen   & ZS      & 50.6 & 47.8 & 27.6 && 53.9 & 98.3 & 65.4 && 2.4 & 100 & 4.8 && 30.6 & 98.1 & 44.9 \\
        \oodrow
        & IntVL  & ZS      & 52.4 & 53.6 & 30.6 && 84.5 & 96.6 & 81.7 && 1.8 & 99.9 & 3.6 && 47.3 & 96.2 & 59.4 \\
        \arrayrulecolor{lightgray}%
        \cmidrule{2-18}
        \arrayrulecolor{black}
        & VMAE   & CMDF    & 83.3 & 85.4 & \bst{83.3} && 91.2 & 99.3 & \bst{93.0} && 92.0 & 98.7 & 90.9 && 95.6 & 98.9 & 96.0 \\
        & VMAE   & Sta     & 82.9 & 85.4 & \sbst{82.9} && 91.8 & 99.1 & \bst{93.0} && 92.6 & 98.9 & \bst{92.0} && 95.8 & 98.9 & \sbst{96.2} \\
        \oodrow
        & VMAE   & Syn     & 45.8 & 50.3 & 34.9 && 61.0 & 96.4 & 66.1 && 27.6 & 96.0 & 34.6 && 52.4 & 93.3 & 60.5 \\
        & VMAE   & St+Sy & 83.0 & 85.4 & \bst{83.3} && 92.6 & 98.9 & \sbst{92.6} && 90.9 & 98.9 & \sbst{91.0} && 96.3 & 98.8 & \bst{96.3} \\
        \midrule
        \oodrow
        \multirow{6}{*}{\rotatebox{90}{UP-Fall~\cite{up-fall}}}
        & Qwen   & ZS      & 44.7 & 31.8 & 20.0 && 21.1 & 90.9 & 27.4 && 0.0 & 100 & 0.0 && 10.5 & 87.4 & 16.6 \\
        \oodrow
        & IntVL  & ZS      & 47.6 & 41.8 & 22.9 && 77.2 & 86.8 & 68.8 && 0.0 & 100 & 0.0 && 38.6 & 81.6 & 47.6 \\
        \arrayrulecolor{lightgray}%
        \cmidrule{2-18}
        \arrayrulecolor{black}
        \oodrow
        & VMAE   & CMDF    & 55.6 & 52.9 & 41.7 && 12.3 & 98.8 & 21.1 && 50.9 & 96.6 & 62.4 && 31.6 & 93.5 & 45.1 \\
        & VMAE   & Sta     & 92.5 & 92.7 & \bst{83.0} && 95.6 & 98.8 & \sbst{95.6} && 87.7 & 97.3 & \bst{88.9} && 92.1 & 94.9 & \bst{92.7} \\
        \oodrow
        & VMAE   & Syn     & 48.0 & 52.9 & 31.3 && 63.2 & 99.3 & 76.2 && 2.6 & 99.5 & 5.0 && 33.8 & 99.0 & 50.0 \\
        & VMAE   & St+Sy & 92.6 & 92.0 & \sbst{82.4} && 96.5 & 99.3 & \bst{96.9} && 79.8 & 98.0 & \sbst{85.4} && 88.2 & 96.3 & \sbst{91.4} \\
        \midrule
        \oodrow
        \multirow{6}{*}{\rotatebox{90}{Le2i~\cite{le2i}}}
        & Qwen   & ZS      & 48.0 & 54.7 & 28.8 && 9.1 & 99.4 & 16.0 && 0.0 & 100 & 0.0 && 4.7 & 99.4 & 8.7 \\
        \oodrow
        & IntVL  & ZS      & 48.0 & 55.2 & 28.1 && 27.3 & 98.3 & 38.7 && 0.0 & 100 & 0.0 && 14.0 & 98.1 & 23.1 \\
        \arrayrulecolor{lightgray}%
        \cmidrule{2-18}
        \arrayrulecolor{black}
        \oodrow
        & VMAE   & CMDF    & 65.9 & 71.9 & 52.7 && 77.3 & 97.8 & 79.1 && 85.7 & 99.5 & \sbst{90.0} && 86.0 & 98.1 & 89.2 \\
        & VMAE   & Sta     & 81.4 & 89.7 & \sbst{72.9} && 100 & 100 & \bst{100} && 100 & 100 & \bst{100} && 100 & 100 & \bst{100} \\
        \oodrow
        & VMAE   & Syn     & 53.1 & 57.1 & 39.5 && 59.1 & 100 & 74.3 && 9.5 & 98.9 & 16.0 && 39.5 & 100 & 56.7 \\
        & VMAE   & St+Sy & 83.2 & 89.7 & \bst{82.4} && 100 & 99.4 & \sbst{97.8} && 100 & 100 & \bst{100} && 100 & 99.4 & \sbst{98.9} \\
        \midrule
        \oodrow
        \multirow{6}{*}{\rotatebox{90}{GMDCSA~\cite{gmdcsa}}}
        & Qwen   & ZS      & 62.4 & 48.4 & 36.1 && 58.8 & 94.7 & 64.5 && 17.6 & 100 & 30.0 && 38.2 & 93.2 & 51.0 \\
        \oodrow
        & IntVL  & ZS      & 55.6 & 44.1 & 33.5 && 58.8 & 92.1 & 60.6 && 0.0 & 100 & 0.0 && 29.4 & 89.8 & 40.0 \\
        \arrayrulecolor{lightgray}%
        \cmidrule{2-18}
        \arrayrulecolor{black}
        \oodrow
        & VMAE   & CMDF    & 59.5 & 54.8 & 50.4 && 64.7 & 89.5 & 61.1 && 52.9 & 100 & 69.2 && 61.8 & 88.1 & 67.7 \\
        & VMAE   & Sta     & 64.5 & 73.1 & \sbst{66.8} && 88.2 & 93.4 & \bst{81.1} && 88.2 & 97.4 & \bst{88.2} && 91.2 & 89.8 & \bst{87.3} \\
        \oodrow
        & VMAE   & Syn     & 64.5 & 60.2 & 50.3 && 52.9 & 97.4 & 64.3 && 23.5 & 100 & 38.1 && 41.2 & 98.3 & 57.1 \\
        & VMAE   & St+Sy & 80.3 & 76.3 & \bst{75.0} && 64.7 & 94.7 & \sbst{68.8} && 82.4 & 97.4 & \sbst{84.8} && 76.5 & 91.5 & \sbst{80.0} \\
        \midrule
        \oodrow
        \multirow{6}{*}{\rotatebox{90}{EDF~\cite{edf-occu}}}
        & Qwen   & ZS      & 51.5 & 38.3 & 20.9 && 0.0 & 100 & 0.0 && 0.0 & 100 & 0.0 && 0.0 & 100 & 0.0 \\
        \oodrow
        & IntVL  & ZS      & 41.5 & 34.4 & 20.7 && 25.0 & 97.3 & 34.8 && 0.0 & 100 & 0.0 && 11.1 & 96.7 & 18.6 \\
        \arrayrulecolor{lightgray}%
        \cmidrule{2-18}
        \arrayrulecolor{black}
        \oodrow
        & VMAE   & CMDF    & 52.6 & 58.6 & 43.7 && 6.2 & 96.4 & 9.5 && 75.0 & 96.3 & \sbst{76.9} && 50.0 & 93.5 & \sbst{60.0} \\
        & VMAE   & Sta     & 56.6 & 75.8 & \bst{57.2} && 82.4 & 95.5 & \bst{77.8} && 100 & 95.3 & \bst{89.4} && 100 & 92.2 & \bst{91.6} \\
        \oodrow
        & VMAE   & Syn     & 33.7 & 43.8 & 24.2 && 35.3 & 92.8 & \sbst{38.7} && 47.6 & 93.5 & 52.6 && 47.4 & 85.6 & 52.2 \\
        & VMAE   & St+Sy & 55.1 & 74.2 & \sbst{54.2} && 82.4 & 95.5 & \bst{77.8} && 100 & 95.3 & \bst{89.4} && 100 & 92.2 & \bst{91.6} \\
        \midrule
        \oodrow
        \multirow{6}{*}{\rotatebox{90}{OCCU~\cite{edf-occu}}}
        & Qwen   & ZS      & 55.1 & 53.5 & 39.5 && 31.2 & 100 & 47.6 && 0.0 & 100 & 0.0 && 15.6 & 100 & 27.0 \\
        \oodrow
        & IntVL  & ZS      & 44.8 & 44.6 & 30.2 && 12.5 & 100 & 22.2 && 0.0 & 100 & 0.0 && 6.2 & 100 & 11.8 \\
        \arrayrulecolor{lightgray}%
        \cmidrule{2-18}
        \arrayrulecolor{black}
        \oodrow
        & VMAE   & CMDF    & 44.2 & 38.6 & 29.4 && 0.0 & 100 & 0.0 && 6.2 & 97.6 & 10.5 && 3.1 & 97.1 & 5.7 \\
        & VMAE   & Sta     & 86.0 & 90.1 & \bst{84.6} && 100 & 97.6 & \bst{94.1} && 100 & 97.6 & \sbst{94.1} && 100 & 94.2 & \sbst{94.1} \\
        \oodrow
        & VMAE   & Syn     & 54.1 & 53.5 & 30.5 && 37.5 & 100 & \sbst{54.5} && 0.0 & 100 & 0.0 && 18.8 & 100 & 31.6 \\
        & VMAE   & St+Sy & 84.4 & 91.1 & \sbst{81.9} && 100 & 97.6 & \bst{94.1} && 100 & 98.8 & \bst{97.0} && 100 & 95.7 & \bst{95.5} \\
        \midrule
        \oodrow
        \multirow{6}{*}{\rotatebox{90}{CaucaF.~\cite{cauca}}}
        & Qwen   & ZS      & 56.3 & 55.3 & 36.2 && 100 & 92.1 & 85.7 && 11.1 & 100 & 20.0 && 55.6 & 89.7 & 64.5 \\
        \oodrow
        & IntVL  & ZS      & 38.1 & 42.6 & 24.5 && 100 & 86.8 & 78.3 && 0.0 & 100 & 0.0 && 50.0 & 82.8 & 56.2 \\
        \arrayrulecolor{lightgray}%
        \cmidrule{2-18}
        \arrayrulecolor{black}
        \oodrow
        & VMAE   & CMDF    & 62.4 & 74.5 & 61.3 && 66.7 & 100 & 80.0 && 100 & 94.7 & 90.0 && 88.9 & 96.6 & 91.4 \\
        & VMAE   & Sta     & 79.5 & 80.9 & \bst{77.5} && 88.9 & 100 & \sbst{94.1} && 100 & 100 & \bst{100} && 94.4 & 100 & \sbst{97.1} \\
        \oodrow
        & VMAE   & Syn     & 60.1 & 55.3 & 40.9 && 77.8 & 94.7 & 77.8 && 22.2 & 100 & 36.4 && 50.0 & 93.1 & 62.1 \\
        & VMAE   & St+Sy & 76.8 & 80.9 & \sbst{76.2} && 100 & 100 & \bst{100} && 100 & 97.4 & \sbst{94.7} && 100 & 96.6 & \bst{97.3} \\
        \bottomrule
      \end{tabular}
      \begin{tablenotes}
        \scriptsize
      \item[\ ] Qwen: Qwen3-VL-8B~\cite{bai_qwen3-vl_2025},
        IntVL: InternVL3.5-8B,
        VMAE: VideoMAE-K400~\cite{tong2022videomae}.
      \item[\ ] ZS: zero-shot,
        CMDF: CMDFall~CS,
        Sta: OF-Sta~CS,
        Syn: OF-Syn~Ra,
        St+Sy: OF-Sta~CS\,+\,OF-Syn~Ra. \quad \colorbox{blue!6}{Shaded} rows: out-of-distribution (test dataset absent from training).
      \end{tablenotes}
    \end{threeparttable}
  }%
\end{table}

\begin{table}[t]
  \centering
  \renewcommand{\arraystretch}{0.88}
  \caption{Evaluation of zero-shot MLLMs and differently fine-tuned VMAE-K400 models on OF-Syn and OF-ItW.
    \textbf{B}al.\ \textbf{Acc}uracy, \textbf{Se}nsitivity,
  \textbf{Sp}ecificity.}
  \label{tab:unified-classification-ood}
  \resizebox{\columnwidth}{!}{%
    \begin{threeparttable}
      \setlength{\tabcolsep}{1.5pt}
      \begin{tabular}{l L L | CCC C CCC C CCC C CCE}
        \toprule
        \multirow{2}{*}{Test} &
        \multirow{2}{*}{Model} &
        \multirow{2}{*}{Train} &
        \multicolumn{3}{c}{16-class} & &
        \multicolumn{3}{c}{Fall $\Delta$} & &
        \multicolumn{3}{c}{Fallen $\Delta$} & &
        \multicolumn{3}{c}{Fall $\cup$ Fallen $\Delta$} \\
        \cmidrule(lr){4-6}\cmidrule(lr){8-10}\cmidrule(lr){12-14}\cmidrule(lr){16-18}
        & & & BAcc & Acc & F1 & &
        Se & Sp & F1 & &
        Se & Sp & F1 & &
        Se & Sp & F1 \\
        \midrule
        \oodrow
        \multirow{6}{*}{\rotatebox{90}{\textbf{OF-Syn}}}
        & Qwen   & ZS      & 55.9 & 67.6 & 48.4 && 54.2 & 96.4 & 59.3 && 15.3 & 98.6 & 23.2 && 43.1 & 95.7 & 53.2 \\
        \oodrow
        & IntVL  & ZS      & 50.3 & 64.9 & 44.3 && 52.4 & 97.2 & 59.8 && 4.7 & 99.5 & 8.4 && 36.5 & 97.3 & 49.2 \\
        \arrayrulecolor{lightgray}%
        \cmidrule{2-18}
        \arrayrulecolor{black}
        \oodrow
        & VMAE   & CMDF    & 32.4 & 51.9 & 31.0 && 33.0 & 96.5 & 40.9 && 10.0 & 97.4 & 14.2 && 27.3 & 94.1 & 35.7 \\
        \oodrow
        & VMAE   & Sta     & 32.0 & 50.9 & 31.2 && 46.2 & 97.3 & 55.1 && 32.7 & 94.1 & 32.2 && 43.1 & 90.9 & 47.3 \\
        & VMAE   & Syn     & 59.0 & 81.4 & \sbst{60.5} && 96.2 & 95.1 & \sbst{81.6} && 67.3 & 97.0 & \bst{66.4} && 90.1 & 92.6 & \bst{81.1} \\
        & VMAE   & St+Sy & 60.5 & 81.2 & \bst{62.4} && 96.2 & 95.8 & \bst{83.6} && 68.7 & 96.7 & \sbst{66.0} && 89.5 & 92.7 & \sbst{81.0} \\
        \midrule
        \oodrow
        \multirow{7}{*}{\rotatebox{90}{\textbf{OF-ItW}}}
        & Qwen   & ZS      & 36.4 & 45.4 & \bst{26.3} && 84.6 & 80.8 & \sbst{67.8} && 19.6 & 98.1 & 29.2 && 69.8 & 78.5 & 66.1 \\
        \oodrow
        & IntVL  & ZS      & 34.9 & 43.8 & \sbst{26.0} && 87.7 & 79.4 & \bst{68.0} && 19.6 & 99.1 & 31.0 && 74.4 & 79.5 & \bst{69.5} \\
        \arrayrulecolor{lightgray}%
        \cmidrule{2-18}
        \arrayrulecolor{black}
        \oodrow
        & VMAE   & CMDF    & 16.0 & 35.3 & 14.7 && 51.1 & 86.9 & 52.3 && 10.0 & 98.7 & 16.6 && 42.9 & 85.7 & 50.3 \\
        \oodrow
        & VMAE   & Sta     & 20.5 & 41.4 & 18.5 && 77.6 & 77.6 & 61.2 && 33.4 & 95.4 & 39.4 && 74.1 & 73.3 & 65.7 \\
        \oodrow
        & VMAE   & Syn     & 20.6 & 46.7 & 21.2 && 72.0 & 84.6 & 64.2 && 35.0 & 96.1 & \bst{42.3} && 69.5 & 81.8 & 67.9 \\
        \oodrow
        & VMAE   & St+Sy & 21.4 & 47.6 & 21.9 && 72.9 & 85.4 & 65.6 && 33.1 & 96.3 & \sbst{41.0} && 68.2 & 82.4 & \sbst{67.5} \\
        \cmidrule{2-18}
        & VMAE   & {\tiny St+Sy+ItW} & 25.8 & 56.5 & 26.7 && 79.3 & 91.3 & 76.0 && 57.2 & 92.8 & 53.6 && 80.9 & 84.7 & 76.9 \\
        \bottomrule
      \end{tabular}
      \begin{tablenotes}
        \scriptsize
      \item[\ ] Qwen: Qwen3-VL-8B~\cite{bai_qwen3-vl_2025},
        IntVL: InternVL3.5-8B,
        VMAE: VideoMAE-K400~\cite{tong2022videomae}.
      \item[\ ] ZS: zero-shot,
        CMDF: CMDFall~CS,
        Sta: OF-Sta~CS,
        Syn: OF-Syn~Ra,
        St+Sy: OF-Sta~CS\,+\,OF-Syn~Ra
        \quad \colorbox{blue!6}{Shaded} rows: out-of-distribution (test dataset absent from training).
      \end{tablenotes}
    \end{threeparttable}
  }%
\end{table}

\paragraph{Fine-Tuning Setup.}
We use VMAE-K400~\cite{tong2022videomae} (87M parameters). As a first probe of out-of-domain generalization, we
fine-tune on individual staged datasets and test each both in-domain
and on OF-ItW (Tab.~\ref{tab:per-dataset-itw}). We then study four
training configurations in more detail:
(1) CMDFall alone (7h), the largest pre-existing staged
dataset, using cross-subject (CS) splits;
(2) OF-Staged (14\,h, 8 datasets) with CS splits;
(3) OF-Synthetic with random (Ra) splits (17h);
(4) OF-Staged\,(CS) + OF-Synthetic\,(Ra) combined.
Single-view durations are comparable (OF-Sta: 13.8\,h, OF-Syn:
16.9\,h), but CS splits leverage multi-view recordings, giving OF-Sta
up to 61.5\,h of training footage.

For metrics, we report accuracy and macro-F1 for the 16-class task,
alongside sensitivity (Se), specificity (Sp), and F1-score for the
critical binary subtasks \emph{Fall} and \emph{Fallen}.
Macro-F1 scores tend to be lower due to the natural class imbalance
across 16 categories.
Tables~\ref{tab:unified-classification}
and~\ref{tab:unified-classification-ood}
provide detailed results; best and second-best values per column are
highlighted in green and orange.

\paragraph{Per-Dataset Generalization.}
Trained on a single staged dataset, a model reaches near-ceiling
in-domain F1 but collapses on OF-ItW
(Tab.~\ref{tab:per-dataset-itw}): Le2i falls from 100 to 58.8 Fall F1,
and Up-Fall from 72.9 to 16.9, with \emph{fallen} detection lower still
(Le2i: 97.7 to 33.1). No staged dataset displays good OOD generalization.

\paragraph{Training on CMDFall.}
We examine CMDFall in more detail, the largest and most diverse staged
dataset (\cref{tab:unified-classification}). On cross-subject splits it
performs strongly in-domain (83.3\% BAcc, Fall F1: 93.0\%), yet still
generalizes only weakly to OF-ItW (Fall F1: 52.3\%, Fallen F1:
16.6\%). Even the largest staged dataset thus lacks the diversity
needed for real-world deployment, particularly for the clinically
critical \emph{fallen} state. We next ask whether unifying datasets
and adding controlled synthetic diversity closes this gap.

\paragraph{Training on OF-Staged.}
Training on the unified OF-Staged superset substantially
improves out-of-distribution performance. Despite similar data volume
to CMDFall alone (61.5h vs 49.8h multi-view), the diversity of eight
datasets with varied environments, subjects, and viewpoints improves
OF-ItW results to 61.2\% Fall F1 and 39.4\% Fallen F1, in-distribution performance remains strong (CMDFall CS BAcc:
82.9\%).
This suggests dataset diversity rather than quantity drives better real-world generalization for both fall events
and fallen states.

\paragraph{Training on OF-Synthetic.}
Training only on synthetic OF-Syn achieves the best single-source
OF-In-the-Wild performance on both Fall F1 (64.2\%) and Fallen F1
(42.3\%), above the OF-Staged superset (61.2\% and 39.4\%). It does so
despite comparable single-view duration (16.9\,h vs.\ 13.8\,h)
and no access to multi-view recordings (OF-Sta CS: up to 61.5\,h). The Fallen F1
improvement reflects OF-Syn's explicit modeling of extended post-fall
ground contact, as discussed in \cref{sec:intro}. This matches
the domain shift analysis (\cref{sec:eval-domain}), where OF-Syn
sits closer to OF-ItW in feature space than staged
datasets.

\paragraph{Training on OF-Staged+OF-Synthetic.}
Combined training on OF-Sta(CS) and OF-Syn(Ra) yields the
best cross-domain Fall F1 on OF-ItW
(Table~\ref{tab:unified-classification-ood}), reaching 65.6\% with 83.0\% BAcc on CMDFall CS.
Improvement over OF-Syn alone is marginal (+1.4 F1 points) suggesting that our synthetic data
captures most of the variation needed for real-world generalization.
\emph{fallen} detection is strongest when
OF-Syn is the sole training source (Fallen F1: 42.3\,\% vs.\
41.0\,\%), possibly due to an underrepresentation of the \emph{fallen} class in staged datasets.

\paragraph{Upper-Bound on OF-ItW.}
To contextualize the cross-domain results, we fine-tune
VMAE-K400 on the combined OF-Staged, OF-Synthetic, and OF-ItW
training splits (Table~\ref{tab:unified-classification-ood}, last
row). This configuration achieves Fall F1 of 76.0 and Fallen F1 of
53.6 (Fall\,$\cup$\,Fallen F1: 76.9). Compared to the best
cross-domain configuration (St+Sy: 65.6 / 41.0 / 67.5), substantial
headroom remains: even a modest amount of
in-the-wild training data improves both \emph{fall} and
\emph{fallen} detection considerably.

\paragraph{Zero-Shot MLLMs.}
As zero-shot baselines, we prompt two 8B-parameter multimodal LLMs
(MLLMs), Qwen3-VL-8B~\cite{bai_qwen3-vl_2025} and InternVL3.5-8B~\cite{wang_internvl35_2025},
to classify video clips into one of the 16 classes without any
task-specific training
(Tables~\ref{tab:unified-classification}
and~\ref{tab:unified-classification-ood}). On OF-ItW, both models
achieve high \emph{fall} F1 scores (67.8 and 68.0, respectively),
yet \emph{fallen} detection remains poor (F1: 29.2 and 31.0),
indicating that detecting sustained post-fall states requires
capabilities that large-scale pre-training alone does not provide.
On staged datasets, both MLLMs score substantially below the
fine-tuned models above. Setup details, prompt templates,
and ablations across additional model scales appear in the appendix.

\paragraph{Fallen Class Confusion.}
Figure~\ref{fig:conf-matrices} illustrates systematic confusion
between semantically adjacent classes in zero-shot MLLM predictions.
Both models classify nearly all \emph{fallen} instances as
\emph{lying} (95--97\% on UP-Fall; $<$3\% correct on CMDFall),
consistent with these being static ground-contact states that likely
require fine-grained temporal reasoning to distinguish. In a care
setting, this confusion would result in missed long-lie alerts, as
the system would label a person unable to get up as merely lying
down. Beyond classification accuracy, 8B-parameter MLLMs are
computationally heavy, which limits their use on edge
devices or for real-time monitoring, unlike the smaller,
task-specific fine-tuned models above.

\paragraph{Fine-Tuned VMAE vs.\ Zero-Shot MLLMs.}
On \emph{fall} event detection, both paradigms perform
comparably on OF-ItW: VMAE (St+Sy, ${\sim}$87M parameters) achieves
Fall F1 of 65.6, close to Qwen3-VL's 67.8 and InternVL3.5's 68.0
(8B parameters each). On the clinically critical \emph{fallen}
state, however, the paradigms diverge: the best fine-tuned
configuration (VMAE trained on OF-Syn) reaches Fallen F1 of 42.3,
substantially outperforming both MLLMs (29.2 and 31.0). This indicates that \emph{fallen}-state recognition
requires task-specific training, a gap that large-scale pre-training
alone does not bridge. At a fraction of the computational cost and model
size, the fine-tuned models are also more suitable for on-site
deployment in care environments.

\paragraph{Summary.}
Across all configurations, \emph{fallen}-state detection is
the primary bottleneck: even the best configuration with in-domain
training data reaches only 53.6\,\% Fallen F1
(Table~\ref{tab:unified-classification-ood}), compared to 76.0\,\%
Fall F1 under the same conditions. Cross-demographic generalization
experiments (cross-age and cross-BMI on OF-Syn) and cross-view results
are in the appendix.

\subsection{Timeline Segmentation}
\label{sec:time-seg}
With dense labels and multi-domain coverage,
OmniFall supports OOD timeline segmentation studies.
To demonstrate, we train ASQuery~\cite{Gan2024ASQueryAQ} at 10\,fps
using frozen SigLIP2~\cite{tschannen2025siglip} features. We consider
three training sets: OF-Staged, OF-Synthetic, and
their union OF-Sta$\cup$OF-Syn, and evaluate on
in-distribution staged test splits and out-of-distribution
OF-ItW. We report F1@IoU (10/25/50), normalized Edit, and
frame accuracy (Table~\ref{tab:segmentation-comparison}).

\paragraph{Training Data Comparison.}
All configurations achieve strong in-distribution performance
(F1@50: 72--78\%), but OF-ItW scores drop substantially.
Training on OF-Synthetic alone achieves
F1@50 of \mbox{27.60}\,\% on OF-ItW, surpassing OF-Staged
(\mbox{20.26}\,\%) and approaching OF-Sta$\cup$OF-Syn
(\mbox{28.61}\,\%). This mirrors the classification results
(Section~\ref{sec:action_classification}): synthetic data with
controlled diversity bridges the domain gap to in-the-wild settings
more effectively than staged data alone.
OF-Sta\,$\cup$\,OF-Syn is the most robust overall: it combines
staged motion structure with synthetic appearance diversity.
\\

Overall, the staged$\rightarrow$wild transfer is hard:
models that score highly in
controlled settings lose precision on OF\mbox{-}ItW, where
appearance, motion, and camera factors vary widely.

\begin{table}[t]
  \centering
  \caption{ASQuery~\cite{Gan2024ASQueryAQ} timeline segmentation with
    SigLIP2 features.}
  \label{tab:segmentation-comparison}
  \setlength{\tabcolsep}{3pt}
  \begin{threeparttable}
  \begin{tabular}{l L | CCC C E}
    \toprule
    \textbf{Test} & \textbf{Train} & \textbf{F1@10} & \textbf{F1@25} & \textbf{F1@50} & \textbf{Edit} & \textbf{Acc} \\
    \midrule
    \multirow{2}{*}{OF-Sta}
      & Sta   & 82.13 & 80.64 & 72.70 & 81.90 & 80.19 \\
      & St+Sy & 82.48 & 80.99 & 73.24 & 82.51 & 79.88 \\
    \midrule
    \multirow{2}{*}{OF-Syn}
      & Syn   & 84.17 & 83.32 & 78.11 & 83.71 & 82.71 \\
      & St+Sy & 83.10 & 82.24 & 76.90 & 82.41 & 81.15 \\
    \midrule
    \oodrow
    \multirow{3}{*}{OF-ItW}
      & Sta   & 39.15 & 33.95 & 20.26 & 36.42 & 43.32 \\
    \oodrow
      & Syn   & 53.92 & 40.56 & 27.60 & 49.97 & 59.39 \\
    \oodrow
      & St+Sy & 52.34 & 45.66 & 28.61 & 48.07 & 59.83 \\
    \bottomrule
  \end{tabular}
  \begin{tablenotes}
    \scriptsize
    \item[\ ] Sta: OF-Sta~CS, Syn: OF-Syn~Ra,
      St+Sy: OF-Sta~CS\,+\,OF-Syn~Ra.
      \quad \colorbox{blue!6}{Shaded} rows: out-of-distribution.
  \end{tablenotes}
  \end{threeparttable}
\end{table}

\section{Risks and Limitations of Synthetic Data Generation}
OF-Synthetic addresses representative gaps concerning demographics and realism of staged fall data.
At the same time, the use of generative video models can introduce new biases. While we manually verified the overall quality of the generation process, we cannot guarantee that all attributes are faithfully representative of global populations, particularly for demographic groups underrepresented in the generator's training data. In order to avoid personal biases regarding ethnicity definitions, we follow the U.S. OMB ethnicity reporting standards~\cite{OMB2024SPD15} which might under-represent certain ethnicities. Furthermore, bound by limitations of the generation model, we limited video generation to a binary gender representation model. We request researchers using our data to be aware of such potential biases and limitations and to consider countermeasures in training and analysis for potential biases of resulting models. Despite these limitations, we believe that our synthetic dataset improves fall detection for all demographic groups.

\section{Conclusion}\label{sec:conc}

\emph{OmniFall} is a unified benchmark spanning
$\sim$80\,hours of densely annotated video across staged, in-the-wild,
and synthetic domains, with framewise 16-class
timeline annotations for both classification and timeline segmentation.

Experiments produce three findings. First, smaller fine-tuned vision
models and zero-shot multimodal LLMs fail differently:
both reach comparable fall-event detection on genuine accidents, yet
diverge on the clinically critical \emph{fallen} state, where
zero-shot models systematically confuse \emph{fallen} with
\emph{lying}. Fine-tuned models trained on synthetic data with
post-fall modeling address this gap at a fraction of the
computational cost, and are better suited to on-site deployment.

Second, training data diversity drives cross-domain generalization
more effectively than volume alone: unifying eight staged datasets
substantially improves in-the-wild performance over the largest
single dataset at comparable training hours, and synthetic data with
controlled diversity improves further. Timeline segmentation
experiments confirm this pattern: synthetic-trained models
outperform staged-only models on out-of-distribution data.

Third, \emph{fallen}-state detection is the main bottleneck:
even with in-domain training data, \emph{fallen} recognition
performs significantly worse than fall-event detection. Since missed \emph{fallen} states
lead to missed long-lie alerts, closing this gap is
the most important challenge for clinical deployment. OmniFall
provides the standardized protocols and multi-domain data to
drive this progress.

\section*{Acknowledgements}
This work has been supported by the Carl Zeiss Foundation through the
JuBot project as well as by funding from the pilot program
Core-Informatics of the Helmholtz Association (HGF).
The authors acknowledge support by the state of Baden-Württemberg through bwHPC.
Experiments were performed on the HoreKa supercomputer funded by the
Ministry of Science, Research and the Arts Baden-Württemberg and by
the Federal Ministry of Education and Research.

\clearpage

\bibliographystyle{splncs04}
\bibliography{bib/fall_datasets,bib/other}
\newpage
\appendix
\section{Appendix}\label{sec:appendix}

This appendix provides supplementary details on the OmniFall
benchmark.
\Cref{sec:app-supplementary} presents the supplementary materials
archive with annotated example videos and an interactive local webpage to explore these videos.
\Cref{sec:app-limitations} discusses limitations and societal impact.
Source dataset provenance and licensing are detailed in
\cref{Sec: Appendix Dataset}, followed by dataset split definitions
(\cref{subsec:cross-subject,subsec:cross-view}), the OF-Synthetic
generation process (\cref{sec:app-ofsyn-gen}), and label taxonomy design
with annotation procedures
(\cref{sec:app-taxonomy,sec:appendix-omnifall-annotation}).
The remaining sections cover setup and ablations for the two
complementary evaluation paradigms of our benchmark: zero-shot
multimodal large language models (MLLMs) applied directly to video
input (\cref{sec:app-mllm-eval}), and smaller trained classifiers
operating on pre-extracted video features
(\cref{sec:app-implementation}).
These paradigms are not directly comparable due to their vastly
different computational footprints, yet both represent highly relevant
directions in current video-based activity recognition research.

\paragraph{Code and Dataset Access.}\label{sec:app-code}
The OmniFall datasets including resources for reproducing our
experiments will be made publicly available upon the acceptance of
our work. This includes the generation code, prompt
templates, hyperparameters, and quality filtering pipeline used to
produce OF-Syn, enabling reproduction and extension of the synthetic
data generation process.

\subsection{Supplementary Materials}\label{sec:app-supplementary}

The accompanying archive contains 168 annotated example videos
spanning all three OmniFall domains: 60 from OF-Synthetic, 48 from
OF-Staged (6 videos from each of the 8 constituent datasets), and 60
from OF-In-the-Wild (evenly sampled fall-containing clips). Each
video is overlaid with
frame-level action labels, timestamps, and a color-coded timeline
progress bar, illustrating the annotation quality and temporal
structure of each domain.

The archive includes an HTML page (\texttt{index\_chrome\_or\_firefox.html}) for browsing
all videos in a web browser. It was tested with Chrome and Firefox.
Videos are encoded in AV1, which is natively supported by these
browsers; standalone video players may require additional codec
installation. Video quality in the archive is reduced (AV1, CRF\,51)
relative to the original recordings to accommodate a larger selection
within the supplementary size limit.

\paragraph{What the examples illustrate.}
The annotated examples allow reviewers to visually verify
several dataset properties referenced in the main paper as well as our manually annotated segmentation labels.
For instance, the unusually high mean segment duration of UP-Fall
(Table~\ref{tab:fall_detection_datasets}, $\bar{t}$\,=\,13.6\,s) is
attributable to extended post-fall lying segments, clearly visible in
the examples. OCCU's defining characteristic, subjects falling behind
furniture, is likewise observable: the resulting occlusion makes
\emph{fallen}-state detection particularly challenging, consistent
with the lower OCCU performance in the classification tables.
OF-Synthetic examples demonstrate the controlled variation across age
groups, environments, and body types described
in~\Cref{sec:app-ofsyn-gen}.

\subsection{Limitations and Societal Impact}\label{sec:app-limitations}

\paragraph{Limitations.} Despite our efforts to create a diverse
benchmark, several limitations remain.
(1) OF-ItW, sourced from web videos of genuine accidents, tends to
overrepresent visually spectacular falls (e.g., from heights, during
sports) relative to the more mundane stumbles and trips that
predominate in elderly care. However, mundane falls are also present
in the dataset, and a robust fall detection system should reliably
detect falls of all severities. In practice, every fall is unique and
does not conform to the controlled scenarios of staged datasets;
cross-domain evaluation remains important for assessing real-world
system performance even if the target distribution introduces a shift
in scenario frequency.
Collecting realistic fall data in actual care facilities is extremely
difficult due to ethical constraints and the infrequency of fall
events, making curated web videos the closest available proxy for
genuine, unscripted accidents.
Our goal extends beyond elderly care to general fall detection
applicable in public health scenarios, workplaces, and recreational
settings, where the diversity of OF-ItW is directly relevant.
(2) OF-Syn's controlled demographic diversity addresses biases in
staged datasets (which predominantly feature young subjects), yet
synthetic data cannot fully capture all real-world complexities and
potentially introduces new,
generator-based biases~\cite{wan2025}.
(3) Deployment in care settings requires careful consideration of
privacy, surveillance ethics, and user acceptance beyond technical
performance.

\paragraph{Societal Impact.} Fall detection systems offer substantial
benefits for health monitoring, potentially saving
lives through timely intervention. Our demonstration that synthetic
data can outperform real staged data while providing controlled
demographic diversity offers a privacy-preserving path forward that
avoids collecting sensitive footage from vulnerable populations.
However, we acknowledge potential negative impacts: surveillance
concerns in care settings, false alarms causing trust erosion,
reduced human oversight, and the risk that synthetic data biases
affect real-world performance across demographics. Responsible
deployment requires transparent performance reporting across
demographic groups, human-in-the-loop decision systems, and ongoing
validation in real care environments.

\subsection{Source Datasets of Omnifall-Staged and Omnifall-In-the-Wild}
\label{Sec: Appendix Dataset}
\paragraph{Search and Identification.}
To collect existing public fall detection datasets from prior work,
we conducted a systematic search derived from PRISMA guidelines
(5)-(9)~\cite{page2021prisma}.
We queried PubMed and Google Scholar using the keywords: [fall] and
[detection] and ([survey] or [review]). 
We removed duplicates, including
pre-prints followed by their peer-reviewed versions. Subsequently, we
conducted an initial manual screening of titles and abstracts to
ensure that the selected surveys were relevant. After this initial
screening, full texts were retrieved and reviewed for eligibility
based on specific inclusion criteria: (1) Surveys with English full
texts. (2) Surveys that have undergone peer-review. (3) Surveys
listing public RGB-based video fall detection datasets. The full PRISMA checklist can be found in the original
publication~\cite{page2021prisma}.

\paragraph{Identified Fall Detection Datasets in OmniFall-Staged.}
Of the 98 surveys retrieved, only 9 met all inclusion criteria. From
these, public datasets (1)-(8) were extracted and included in
OF-Staged are listed in Table \ref{tab:fall_detection_datasets} in
the main paper and listed in the following paragraph.

\paragraph{Source Datasets.}

The benchmark unifies eight (OF-Staged) existing fall detection
datasets, each accessed under appropriate permissions or licenses and OF-ItW makes use of videos from the OOPS dataset:

\begin{itemize}
  \item \textbf{CMDFall}~\cite{cmdfall}: Used with explicit
    permission from the authors.\\
    \href{https://www.mica.edu.vn/perso/Tran-Thi-Thanh-Hai/CMDFALL.html}{https://www.mica.edu.vn/perso/Tran-Thi-Thanh-Hai/CMDFALL.html}

  \item \textbf{UP-Fall}~\cite{up-fall}: Used with explicit
    permission from the authors.\\
    \href{https://sites.google.com/up.edu.mx/har-up/}{https://sites.google.com/up.edu.mx/har-up/}

  \item \textbf{Le2i}~\cite{le2i}: Available under CC-BY-NC-SA 3.0 license.\\
    \href{https://search-data.ubfc.fr/imvia/FR-13002091000019-2024-04-09_Fall-Detection-Dataset.html}{https://search-data.ubfc.fr/imvia/FR-13002091000019-2024-04-09\_Fall-Detection-Dataset.html}

  \item \textbf{GMDCSA24}~\cite{gmdcsa}: Available under MIT license.\\
    \href{https://github.com/ekramalam/GMDCSA24-A-Dataset-for-Human-Fall-Detection-in-Videos}{https://github.com/ekramalam/GMDCSA24-A-Dataset-for-Human-Fall-Detection-in-Videos}

  \item \textbf{EDF and OCCU}~\cite{edf-occu}: Used with explicit
    permission from the authors.\\
    \href{https://zenodo.org/records/15494102}{https://zenodo.org/records/15494102}

  \item \textbf{CaucaFall}~\cite{cauca}: Available under CC BY 4.0 license.\\
    \href{https://data.mendeley.com/datasets/7w7fccy7ky/4}{https://data.mendeley.com/datasets/7w7fccy7ky/4}

  \item \textbf{MCFD}~\cite{mcfd}: Used with explicit permission from
    the authors.\\
    \href{https://www.iro.umontreal.ca/~labimage/Dataset/}{https://www.iro.umontreal.ca/~labimage/Dataset/}

  \item \textbf{OOPS}~\cite{oops}: Available under CC BY-NC-SA 4.0 license.\\
    \href{https://oops.cs.columbia.edu/data/}{https://oops.cs.columbia.edu/data/}
\end{itemize}

\subsection{Dataset Splits.}
\subsubsection{OF-Staged}
\textbf{OF-Staged} spans 14\,h of recordings (only counting single-view duration) across eight datasets
(~\cref{fig:dataset-sizes}). Here we list the
individual per dataset CS/CV splits which are also used in the
combined OF-Staged CS/CV split. Note that the main paper only shows CS results since the CS split leverages all possible views and therefore contains significantly more training data, while results for the CV split are shown in \cref{tab:unified-classification-cv}.

\paragraph{Cross-Subject Splits.}
\label{subsec:cross-subject}
We implement dataset-specific subject splits, following original
protocols where available: CMDFall (following published odd/even
  protocol: train: odd-IDs 1-49, val: IDs 2/8/18/42/48, test: remaining
20 even-IDs); CaucaFall (train: S1-S7, val: S10, test: S8-S9, with
S6/S8 being nighttime recordings); EDF (train: Jianjun/Jinhui/Peter,
val: Songsong, test: Zhong); OCCU (train: Jiayan/Wangde/Yangyi, val:
Zhangbo, test: Zhouhui); GMDCSA (train: subjects 1/3, val: subject 2,
test: subject 4); Le2i (train: subjects 0/1/3/4/6/8, val: subject 5,
test: subjects 2/7); UP Fall (train: IDs 1/3/5-10/12-14/17, val: ID
11, test: IDs 2/4/15/16). MCFD's single subject is used only for
training in cross-subject evaluation.

\paragraph{Cross-View Splits.}
\label{subsec:cross-view}
For multi-view datasets, we create camera-based splits: CMDFall
(train: camera 1, val: camera 5, test: cameras 2/3/4/6); EDF and OCCU
(each using train/val: view 1, test: view 2); GMDCSA (train: subjects
  2/3 location with identical room/sensor, val: subject 1 location,
test: subject 4 location); Le2i (train: Coffee-room and Lecture-room
views, val: Office view, test: Home-room views); MCFD (train: camera
1, val: camera 3, test: cameras 2/4/5/6/7/8); UP Fall (train/val:
camera 2, test: camera 1). Single-view CaucaFall participates only in
training for cross-view evaluation.

\subsubsection{OF-Synthetic}
OF-Synthetic provides four evaluation protocols designed to probe
demographic and phenotypic generalization across age, ethnicity, and
body composition.

\paragraph{Random Split (80/10/10).}
A standard baseline dividing the dataset into 80\% training, 10\%
validation, and 10\% testing, comprising 9,600/1,200/1,200 videos
with corresponding 15,344/1,956/1,928 action segments. This split is
used by default in our experiments.

\paragraph{Cross-Age Split.}
This split evaluates transfer across age groups by training on adults
(young adults and middle-aged individuals), validating on teenagers,
and testing on children, toddlers, and elderly individuals. The
configuration comprises 4,000 training videos, 2,000 validation
videos, and 6,000 test videos.

\paragraph{Cross-Ethnicity Split.}
To assess generalization across large phenotypic variations, training
includes individuals identified as \emph{White}, \emph{Asian}, and
\emph{Hispanic or Latino}; validation uses \emph{American Indian or
Alaska Native}; and testing targets \emph{Black or African American},
\emph{Middle Eastern or North African}, and \emph{Native Hawaiian or
Pacific Islander}. The split sizes are 5,178 training, 1,741
validation, and 5,081 test videos.

\paragraph{Cross-BMI Split.}
This protocol examines robustness to body-type variation by training
on underweight and normal-weight individuals, validating on
overweight individuals, and testing on obese individuals, with
6,066/2,962/2,972 videos per split.

\subsection{OF-Synthetic Data Generation Process}
\label{sec:app-ofsyn-gen}
Existing fall detection datasets exhibit systematic demographic
biases, predominantly featuring young to middle-aged participants
with limited ethnic diversity. To address these limitations while
avoiding ethical complexities of collecting fall data from vulnerable
elderly populations, we generated OF-Synthetic with explicit control
over demographic attributes through structured prompt design for the
Wan 2.2 video diffusion model, ensuring systematic coverage across
age groups, ethnicities, body types, and environmental contexts.

\subsubsection{Demographic Attribute Design}

\paragraph{Age Groups.}
Six age categories follow common reporting practices: toddlers (1--4
years), children (5--12 years), teenagers (13--17 years), young
adults (18--34 years), middle-aged (35--64 years), and elderly (65+
years). Each age group receives equal
representation with 200 generated videos per action class.

\paragraph{Race and Ethnicity.}
To ensure systematic ethnic diversity, we adopt the seven categories
defined by the U.S. Office of Management and Budget (OMB) standard
for race and ethnicity reporting~\cite{OMB2024SPD15}: Black, Asian,
White, Hispanic/Latino, Middle Eastern/North African (MENA), American
Indian/Alaska Native (AIAN), and Native Hawaiian/Pacific Islander
(NHPI). This standardized taxonomy enables consistent demographic
reporting and comparison with population statistics.

\paragraph{Skin Tone.}
Beyond categorical ethnicity labels, we incorporate the Monk Skin
Tone Scale~\cite{monk2023monk}, a perceptually uniform 10-point scale
(MST1--MST10) ranging from very light to completely dark skin. This
continuous representation captures intra-ethnic skin tone variation
and provides finer-grained control over visual appearance diversity
than ethnicity categories alone.

\paragraph{Body Morphology.}
Four BMI bands (underweight, normal, overweight, obese) and three
height categories (short, average, tall) provide systematic body type
coverage. BMI bands translate to appearance descriptors (thin,
average build, chubby, stocky) in prompts. Gender presentation is
specified as male or female.

Beyond person-level attributes, environmental context and scenario
characteristics provide additional sources of controlled variation.

\subsubsection{Environmental and Contextual Diversity}

\paragraph{Age-Appropriate Environments.}
Each age group is associated with 100 indoor and 100 outdoor
environment descriptors reflecting typical activity contexts for that
demographic.

Younger age groups feature educational and play-oriented settings:
toddler environments include daycare play areas, indoor playgrounds,
and sandbox areas; children's environments span elementary
classrooms, playgrounds, and sports fields; teenage environments
encompass high school hallways, athletic facilities, and teen
activity centers. Adult age groups reflect professional and
residential contexts: young adult environments include offices, gyms,
universities, and apartments; middle-aged environments feature
corporate settings, professional spaces, and home environments;
elderly environments comprise senior centers, assisted living
facilities, medical clinics, and residential areas. This age-specific
environment mapping ensures generated scenarios reflect realistic
activity contexts for each demographic.

\paragraph{Clothing Diversity.}
Each age group has 200+ age-appropriate clothing options spanning
everyday casual wear (60\%), seasonal attire (25\%), sleepwear
(10\%), and special occasion clothing (5\%). Examples range from
toddlers' colorful casual wear to elderly soft comfortable clothing,
ensuring contextually appropriate attire with visual diversity.

\subsubsection{Action-Specific Attributes}

Beyond demographic and environmental attributes, which remain
constant across action classes, each action class incorporates
action-specific descriptors to control temporal dynamics and scenario
characteristics.

\paragraph{Fall-Specific Attributes.}
Fall videos include 30 fall mechanism types (slip on wet surface,
trip over obstacle, lose balance from dizziness), 20 fall direction
specifications (falling forward, backward, sideways, diagonally,
while spinning), 20 recovery attempt descriptions (reaching for
support, windmilling arms, bracing for impact), age-appropriate prior
activities, and 20 facial expression descriptors.

The inclusion of facial expressions addresses a specific challenge
with the Wan 2.2 model: the model exhibits a tendency to generate
individuals with smiling or happy expressions, presumably due to
training data distribution or alignment objectives. This creates
inappropriate and unrealistic scenarios where individuals smile while
falling. To counteract this bias, we explicitly specify contextually
appropriate facial expressions (startled, panicked, focused,
surprised) in fall scenario prompts.

\paragraph{Other Action Classes.}
Similar attribute sets were defined for other action classes (lie
down, stand up, sit down, walk), incorporating action-specific motion
descriptors, environmental constraints, and contextual details
appropriate to each activity. The ``other'' category receives special
attention to ensure minimal overlap with the 15 primary action
classes through pre-screening of candidate activities for distinctiveness.

\subsubsection{Camera Specification}

To complete the prompt specification, camera parameters control
viewpoint variation. Each video specifies camera vantage point
through four attributes: elevation (eye-level, low, high/CCTV
perspective, top-down/birds eye), azimuth (front, rear, left, right),
distance (medium/distant shot, far/long-distance shot), and shot type
(static wide, static medium wide). Frame rates are specified as
24fps, 25fps, or 30fps real-time. This systematic camera parameter
variation produces videos from diverse viewpoints representative of
real-world surveillance and monitoring camera placements.

\subsubsection{Prompt Construction and Generation}

\paragraph{Prompt Structure.}
Each prompt combines demographic, environmental, action-specific, and
camera attributes into a structured natural language description. A
typical fall prompt follows this template:

\begin{quote}
  \textit{``A [height] [body type] [age descriptor] [gender] with
    [skin tone] and [ethnicity] features wearing [clothing] is in the
    process of falling due to [fall mechanism], [fall direction], in
    [environment], while [recovery attempt], after [prior activity],
    with [facial expression], [framerate], [shot type], camera
    positioned [elevation] from the [azimuth], it is a [distance].
  Realistic. Fully dressed upper body.''}
\end{quote}

The modifiers ``Realistic'' and ``Fully dressed upper body'' were
found beneficial for generating appropriate video content.

\paragraph{Dataset Scale.}
We generated 200 videos per age group for 10 base action classes
(fall, fallen, lie down, lying, sit down, sitting, stand up,
standing, walk, other), yielding 200 $\times$ 6 age groups $\times$
10 actions = 12,000 videos totaling 17 hours of synthetic footage at
1280$\times$720 resolution and 16fps. While the initial generation
targeted these 10 action classes, subsequent dense annotation
revealed the utility of additional fine-grained distinctions, leading
to the expansion to the full 16-class taxonomy used for timeline
segmentation evaluation.

\subsubsection{Safety Filtering and Iterative Refinement}

The Wan 2.2 model occasionally produced inappropriate content
requiring iterative filtering: generated videos were manually
reviewed, inappropriate content removed, and prompts refined to
reduce recurrence. This process repeated until the dataset contained
only safe-for-work content suitable for public release.

\subsection{OmniFall Dense Annotation Process}
\label{sec:appendix-omnifall-annotation}

Unifying eight staged datasets, curating genuine accident videos, and
annotating 12,000 synthetic videos into a consistent 16-class
taxonomy required substantial annotation effort. This section details
the annotation workflow, software tooling, and quality assurance
measures employed to ensure consistent, frame-level labels across the
33 hours of single-view footage (82 hours including multi-view
recordings).

\subsubsection{Taxonomy Design and Rationale}
\label{sec:app-taxonomy}

The 16-class taxonomy was designed to satisfy two requirements:
clinical relevance for fall-related care and sufficient granularity
to cover the observable activities across all source datasets.

\paragraph{Clinical motivation.}
Interviews with medical professionals identified the distinction
between the transient \emph{fall} event and the sustained
\emph{fallen} state as clinically essential. Prolonged ground contact
after a fall (``long-lie'' episodes) substantially increases
mortality risk~\cite{tinetti1993predictors,fleming2008inability}, yet
most existing datasets annotate only a binary fall/no-fall label or
treat the entire fall sequence as a single event. To enable detection
of long-lie situations even when the initial fall event is missed, the
taxonomy explicitly separates transient actions from their resulting
static states.

\paragraph{Action-state pairing principle.}
This separation generalizes beyond falls. Four transient actions
(\emph{fall}, \emph{sit\_down}, \emph{lie\_down}, \emph{stand\_up})
are each paired with corresponding static states (\emph{fallen},
\emph{sitting}, \emph{lying}, \emph{standing}), enabling models to
distinguish ongoing activities from completed postures. The class
\emph{walk} captures locomotion, while \emph{other} serves as a
catch-all for activities not covered by the remaining classes.

Two pairs are particularly critical for practical deployment.
First, \emph{fallen} must be distinguished from \emph{lying}: both
involve ground contact, yet only the former signals an emergency. A
system that cannot separate them will either miss genuine post-fall
states or trigger continuous false alarms whenever a person lies down
intentionally, rendering it unusable in environments where lying is
routine (e.g., yoga, rest areas, bedrooms). Second, \emph{fall} must
be distinguished from \emph{lie\_down}: both involve a transition
toward the ground, but only an uncontrolled descent warrants an
alert. Most existing datasets collapse these pairs into a single
class or omit them entirely, preventing models from learning the
distinction. Our experimental results confirm that this confusability
is the primary bottleneck: even large-scale zero-shot MLLMs
systematically classify \emph{fallen} as \emph{lying}
, as highlighted by the zero-shot confusion matrices in
Figure~\ref{fig:conf-matrices} where both MLLMs classify nearly all
\emph{fallen} instances as \emph{lying}. Across all training
configurations, \emph{fallen} detection remains the weakest point.

\paragraph{Extended classes.}
Six additional classes (\emph{kneel\_down}, \emph{kneeling},
\emph{squat\_down}, \emph{squatting}, \emph{crawl}, \emph{jump})
emerged during dense annotation of OF-Synthetic and OF-ItW videos,
where recognizable instances of these activities were observed.
They follow the same action-state pairing principle and are retained
for completeness, although they occur infrequently in staged datasets
and can be mapped to \emph{other} (class 9) for compatibility.

\paragraph{Class definitions.}
Table~\ref{tab:label-definitions} provides the precise definition
used for each class during annotation. Key distinctions include:
\emph{fall} denotes any uncontrolled descent (including onto a bed,
if not a controlled lying-down with arm support), while
\emph{lie\_down} is intentional; \emph{fallen} begins when
inertia-driven movement from the fall stops, and may include
ground-level movement unrelated to fall or recovery;
\emph{stand\_up} also covers transitions from lying to sitting.

\begin{table}[t]
  \centering
  \small
  \caption{OmniFall 16-class label definitions. Core classes (0--9)
    are present in all datasets; extended classes (10--15) appear in
  OF-ItW and OF-Syn only.}
  \label{tab:label-definitions}
  \setlength{\tabcolsep}{3pt}
  \begin{tabular}{clp{6.5cm}}
    \toprule
    \textbf{ID} & \textbf{Label} & \textbf{Definition} \\
    \midrule
    0  & walk       & Locomotion including jogging and running. Not
    when pushing a large object; includes carrying small items. \\
    1  & fall       & Uncontrolled descent from any prior state.
    Includes falling onto a bed if the motion is not controlled. \\
    2  & fallen     & On the ground or mattress after a fall. Begins
    when inertia-driven motion stops. \\
    3  & sit\_down  & Controlled transition to a seated position on a
    chair, bed, or ground. \\
    4  & sitting    & Resting in a seated position. \\
    5  & lie\_down  & Intentional transition to a lying position (in
    contrast to \emph{fall}). \\
    6  & lying      & Resting in a lying position after intentionally
    getting into that position. \\
    7  & stand\_up  & Rising from fallen, lying, or sitting. Also
    covers lying-to-sitting transitions. \\
    8  & standing   & Stationary upright posture without walking. \\
    9  & other      & Any activity not covered by the remaining
    classes. \\
    \midrule
    10 & kneel\_down  & Transition to a kneeling position. \\
    11 & kneeling     & Static kneeling posture. \\
    12 & squat\_down  & Transition to a squatting position. \\
    13 & squatting    & Static squatting posture. \\
    14 & crawl        & Locomotion on hands and knees. \\
    15 & jump         & Jumping, including from elevated positions. \\
    \bottomrule
  \end{tabular}
\end{table}

\paragraph{Temporal boundary conventions.}
Transient actions are annotated starting from the first frame where
the motion visibly diverges from the preceding activity. Static states
begin at the first frame where the person reaches a resting posture.
Natural label sequences such as \emph{fall}, \emph{fallen},
\emph{stand\_up} do not always appear together: a person may rise
immediately after falling without a \emph{fallen} segment, or sit
down and stand up without a \emph{sitting} segment. The distinction
between \emph{sit\_down} followed by \emph{lie\_down} versus a
single \emph{lie\_down} depends on whether a moment of rest in the
sitting position is observable.

\paragraph{Why complete re-annotation.}
Source datasets use incompatible label schemes ranging from binary
fall/no-fall to coarse ADL categories, with inconsistent temporal
granularity and label semantics. All videos were re-annotated from scratch using the
unified definitions in Table~\ref{tab:label-definitions}. This
ensures consistent label semantics and complete frame-level temporal
coverage across all datasets, rather than propagating inconsistencies
from differing annotation conventions.

\subsubsection{Annotation Software Selection and Customization}

After systematic comparison, we selected VGG Image Annotator
(VIA)~\cite{dutta2019vgg} as the most efficient platform among
available open-source video annotation tools. Custom modifications
introduced comprehensive keyboard-only annotation capabilities,
enabling all operations without mouse interaction. This accelerated
annotation throughput through rapid temporal navigation, label
assignment, and boundary adjustment via keyboard shortcuts,
particularly beneficial for precise frame-level timeline segmentation.

\subsubsection{Annotation Team and Training}

Ten annotators received detailed instruction on the
16-class taxonomy, emphasizing visually similar action pairs (e.g.,
\emph{fall} vs.\ \emph{lie down}) and temporal boundaries between
transient actions and static states. Frequent team discussions
established consensus on challenging cases, including transition
frame determination (\emph{fall} to \emph{fallen}), controlled vs.
uncontrolled descents, and handling occluded or out-of-frame actions.

\subsubsection{Quality Assurance Protocol}

Nearly all videos underwent two-pass review: initial frame-level
annotation followed by quality control verifying temporal boundary
accuracy, label correctness, and taxonomy consistency. Discrepancies
were corrected, with ambiguous cases referred to team discussion.

\subsection{Zero-Shot MLLM Setup and Ablations}\label{sec:app-mllm-eval}

MLLMs are highly susceptible to variations in prompt design and
input/output formatting. Rather than arbitrarily selecting a
configuration, we provide detailed ablations over the exact prompt
setup to document the best achievable zero-shot performance for each
model. This is particularly important because even minor changes to
individual prompt components can shift fall detection F1 by up to 9
percentage points on the same model and data
(Table~\ref{tab:component-ablation}).

\begingroup
\renewcommand{\arraystretch}{1.2}
\begin{table}[t]
  \centering
  \caption{\textbf{Zero-shot MLLM scale comparison} on OF-ItW
    for models up to 8B parameters.
    \textbf{B}alanced \textbf{Acc}uracy,
    \textbf{Se}nsitivity, \textbf{Sp}ecificity.
  Best results \textbf{bold}, second-best \underline{underlined}.}
  \label{tab:zeroshot-scale}

  \begin{tabular}{@{}l rrr rrr rrr@{}}
    \toprule
    \multirow{2}{*}{\textbf{Model}} &
    \multicolumn{3}{c}{16-class} &
    \multicolumn{3}{c}{Fall $\Delta$} &
    \multicolumn{3}{c}{Fallen $\Delta$} \\
    \cmidrule(lr){2-4} \cmidrule(lr){5-7} \cmidrule(lr){8-10}
    & \multicolumn{1}{c}{BAcc} & \multicolumn{1}{c}{Acc} & \multicolumn{1}{c}{F1}
    & \multicolumn{1}{c}{Se}   & \multicolumn{1}{c}{Sp}  & \multicolumn{1}{c}{F1}
    & \multicolumn{1}{c}{Se}   & \multicolumn{1}{c}{Sp}  & \multicolumn{1}{c}{F1} \\
    \midrule
    InternVL3.5-2B~\cite{wang_internvl35_2025}
    & 34.8 & 41.6 & 24.4
    & 76.9 & 82.3 & 64.9
    & \textbf{36.0} & 94.5 & \textbf{40.2} \\
    Qwen3-VL-2B~\cite{bai_qwen3-vl_2025}
    & 28.2 & 29.6 & 15.7
    & 75.5 & 79.4 & 61.5
    & 4.8 & 98.5 & 8.3 \\
    \addlinespace
    InternVL3.5-4B
    & 28.3 & 36.9 & 20.9
    & 71.6 & \textbf{86.0} & 65.4
    & 7.1 & 95.7 & 10.0 \\
    Qwen3-VL-4B
    & \textbf{38.1} & 41.7 & 24.3
    & 80.1 & \underline{84.5} & \textbf{68.9}
    & 5.8 & \textbf{99.5} & 10.6 \\
    \addlinespace
    InternVL3.5-8B
    & 34.9 & \underline{43.8} & \underline{26.0}
    & \textbf{87.7} & 79.4 & \underline{68.0}
    & 19.6 & \underline{99.1} & 31.0 \\
    Qwen3-VL-8B
    & 36.4 & \textbf{45.4} & \textbf{26.3}
    & \underline{84.6} & 80.8 & 67.8
    & 19.6 & 98.1 & 29.2 \\
    Keye-VL-1.5-8B~\cite{yang_kwai_2025}
    & \underline{36.5} & 40.6 & 24.1
    & 78.0 & 84.0 & 67.2
    & \underline{26.7} & 98.7 & \underline{39.1} \\
    \bottomrule
  \end{tabular}
\end{table}
\endgroup

\paragraph{Zero-Shot MLLM Setup.}\label{sec:app-mllm}
We evaluate Qwen3-VL~\cite{bai_qwen3-vl_2025},
InternVL3.5~\cite{wang_internvl35_2025}, and
Keye-VL-1.5~\cite{yang_kwai_2025} in a zero-shot setting using
vLLM~\cite{kwon_efficient_2023} with PagedAttention for inference.
Greedy decoding is used with temperature~0.0, top-$p$~0.8,
top-$k$~20, a fixed seed of~0, and a maximum of 1024 new tokens.
Each annotated temporal segment is sampled as a 2-second clip of
$T{=}16$ frames at 7.5\,FPS. Segments shorter than two seconds are
padded with adjacent context; longer segments are subsampled via a
random offset with fixed seed. Spatially, the shortest edge is
resized to 448 pixels and center-cropped to a $448 \times 448$ square,
yielding a $16 \times 448 \times 448 \times 3$ input tensor per clip.
Model responses are parsed with keyword-based regex that tolerates
minor deviations (\eg ``walking'' maps to \texttt{walk});
unrecognized outputs are mapped to \texttt{other}.

Listing~\ref{lst:zeroshot-prompt} shows the final prompt template.
The \texttt{<video>} token represents the visual input, and
\texttt{\{ALLOWED\_LABELS\}} is expanded to the 16 taxonomy classes
defined in Table~\ref{tab:label-definitions}. The prompt uses
free-text output with a fixed trigger phrase rather than structured
JSON, and includes a role prompt without class definitions, as
determined by the ablation below.

\begin{lstlisting}[
  caption={Zero-shot prompt template for MLLM action classification.},
  label=lst:zeroshot-prompt,float=tp]
<video>
Role:
You are an expert Human Activity Recognition
(HAR) specialist.

Task:
Analyze the provided video clip and classify
the primary action being performed.
Only use one of the following allowed labels.

{ALLOWED_LABELS}

Output Format:
Respond with: "The best answer is: <class_label>"
where <class_label> is one of the allowed labels.
\end{lstlisting}

\paragraph{Prompt Design Ablation.}
Table~\ref{tab:zeroshot-prompt-ablation} ablates three prompt design
choices on OF-ItW for the two 8B models used in the main evaluation:
output format (free text \vs JSON), role prompting, and class
definitions. Free-text generation consistently outperforms
JSON-structured output for InternVL3.5-8B, where JSON
reduces balanced accuracy by up to 7 points; for Qwen3-VL-8B the
difference between formats is negligible. Role prompting improves
balanced accuracy for both models. Including class definitions
improves macro F1 slightly, but reduces balanced accuracy, indicating
a bias toward defined classes. The final prompt therefore uses
free-text output with role prompting and no definitions.

\begin{table}[t]
  \centering
  \caption{\textbf{Prompt design ablation} on OF-ItW (16-class).
    Effect of output format (free text \vs JSON), role prompting, and
    class definitions on balanced accuracy and macro F1. Best per
  model \textbf{bold}, second-best \underline{underlined}.}
  \label{tab:zeroshot-prompt-ablation}
  \begin{tabular}{@{}c c c rr rr@{}}
    \toprule
    \multirow{2}{*}{\textbf{Output}} &
    \multicolumn{2}{c}{\textbf{Components}} &
    \multicolumn{2}{c}{\textbf{InternVL3.5-8B}} &
    \multicolumn{2}{c}{\textbf{Qwen3-VL-8B}} \\
    \cmidrule(lr){2-3} \cmidrule(lr){4-5} \cmidrule(lr){6-7}
    & Role & Def. & BAcc & MF1 & BAcc & MF1 \\
    \midrule
    \multirow{4}{*}{Text}
    & \ding{55} & \ding{55} & 34.3 & 25.2 & 34.1 & 25.2 \\
    & \ding{51} & \ding{55} & \textbf{34.9} & 26.0
    & \textbf{36.4} & 26.3 \\
    & \ding{55} & \ding{51} & 31.5 & 26.5 & 34.1 & \textbf{27.1} \\
    & \ding{51} & \ding{51} & 32.3 & \textbf{26.9}
    & 35.3 & 27.0 \\
    \midrule
    \multirow{4}{*}{JSON}
    & \ding{55} & \ding{55} & 27.3 & 22.5 & 34.2 & 25.6 \\
    & \ding{51} & \ding{55} & 29.7 & 24.6 & 36.4 & 26.7 \\
    & \ding{55} & \ding{51} & 26.9 & 23.1 & 32.7 & 26.2 \\
    & \ding{51} & \ding{51} & 27.3 & 23.1 & 34.2 & 26.7 \\
    \bottomrule
  \end{tabular}
\end{table}

\paragraph{Role Prompt Location.}
Table~\ref{tab:role-location} compares placing the role prompt as a
system message versus embedding it in the user message for Qwen3-VL-8B
on OF-ItW. Placing the role in the user message, where it appears
after the visual tokens, yields higher balanced accuracy (36.1
vs.\ 34.1) and macro F1 (26.0 vs.\ 25.1). All subsequent experiments
therefore use user-message placement.

\begin{table}[t]
  \centering
  \caption{\textbf{Role prompt location ablation} for Qwen3-VL-8B on
    OF-ItW. ``System'' places the role as a system message (before
    visual tokens), ``User'' embeds it in the user message (after
  visual tokens).}
  \label{tab:role-location}
  \begin{tabular}{@{} crr @{}}
    \toprule
    \textbf{Location} & \textbf{BAcc} & \textbf{MF1} \\
    \midrule
    System & 34.1 & 25.1 \\
    User   & 36.1 & 26.0 \\
    \bottomrule
  \end{tabular}
\end{table}

\paragraph{Prompt Component Sensitivity.}
Table~\ref{tab:component-ablation} provides a fine-grained
one-at-a-time ablation over four prompt components for Qwen3-VL-8B on
OF-ItW, with output format fixed to free text. Starting from a
minimal baseline (marked $\ast$), each section varies one dimension
while keeping all others at their baseline setting. For the
\emph{role} component, a standard HAR specialist persona yields the
best balanced accuracy (36.1) and fall F1 (68.1), while specialized
fall-detection personas improve fallen detection (F1 up to 45.4) at
the cost of overall accuracy. Varying the \emph{task instruction} from
minimal to extended phrasing increases fallen sensitivity (30.5
vs.\ 13.8) but reduces balanced accuracy by over 5 points. Among
\emph{label formats}, the grouped format achieves the highest balanced
accuracy (35.6) and fall specificity (81.2), whereas bulleted and
comma-separated formats perform comparably on most metrics. Adding
class \emph{definitions} consistently improves fallen detection (F1
from 21.6 to 44.7--45.2) but does not improve overall balanced
accuracy, confirming the bias toward defined classes observed in the
coarse ablation above.

\begingroup
\renewcommand{\arraystretch}{1.2}
\begin{table}[t]
  \centering
    \begin{tabular}{@{}l l rrr rrr rrr@{}}
      \toprule
      \multirow{2}{*}{Component} & \multirow{2}{*}{Variant} &
      \multicolumn{3}{c}{16-class} &
      \multicolumn{3}{c}{Fall $\Delta$} &
      \multicolumn{3}{c}{Fallen $\Delta$} \\
      \cmidrule(lr){3-5} \cmidrule(lr){6-8} \cmidrule(lr){9-11}
      & & \multicolumn{1}{c}{BAcc} & \multicolumn{1}{c}{Acc}
      & \multicolumn{1}{c}{F1} & \multicolumn{1}{c}{Se}
      & \multicolumn{1}{c}{Sp} & \multicolumn{1}{c}{F1}
      & \multicolumn{1}{c}{Se} & \multicolumn{1}{c}{Sp}
      & \multicolumn{1}{c}{F1} \\
      \midrule
      \multirow{4}{*}{Role}
      & none$^{\ast}$    & 33.6 & 42.9 & 24.7 & 85.2 & 75.9 & 63.8 &
      13.8 & 98.1 & 21.6 \\
      & standard         & \textbf{36.1} & 45.2 & 26.0 & 85.7 &
      \textbf{80.7} & \textbf{68.1} & 18.9 & \textbf{98.2} & 28.4 \\
      & specialized      & 32.5 & \textbf{45.9} & \textbf{28.8} &
      92.8 & 64.8 & 59.3 & 35.5 & 96.6 & 43.9 \\
      & video spec.      & 30.3 & \textbf{45.9} & 26.9 &
      \textbf{94.2} & 63.9 & 59.4 & \textbf{36.8} & 96.8 & \textbf{45.4} \\
      \midrule
      \multirow{2}{*}{Task}
      & standard$^{\ast}$ & \textbf{33.6} & \textbf{42.9} & 24.7 &
      85.2 & \textbf{75.9} & \textbf{63.8} & 13.8 & \textbf{98.1} & 21.6 \\
      & extended          & 28.3 & 42.7 & \textbf{25.0} &
      \textbf{95.4} & 56.4 & 55.5 & \textbf{30.5} & 97.1 & \textbf{39.9} \\
      \midrule
      \multirow{4}{*}{Labels}
      & bulleted$^{\ast}$ & 33.6 & \textbf{42.9} & \textbf{24.7} &
      85.2 & 75.9 & 63.8 & 13.8 & \textbf{98.1} & 21.6 \\
      & numbered          & 31.1 & 39.0 & 23.3 & 74.2 & 80.1 & 61.3 &
      \textbf{25.8} & 95.3 & \textbf{31.7} \\
      & grouped           & \textbf{35.6} & 42.4 & \textbf{24.7} &
      79.9 & \textbf{81.2} & \textbf{65.5} & 12.6 & 97.9 & 19.5 \\
      & comma             & 35.1 & 42.2 & 24.5 & \textbf{85.7} & 75.1
      & 63.4 & 16.0 & 97.4 & 23.5 \\
      \midrule
      \multirow{3}{*}{Definitions}
      & none$^{\ast}$  & 33.6 & 42.9 & 24.7 & 85.2 & 75.9 & 63.8 &
      13.8 & \textbf{98.1} & 21.6 \\
      & standard       & \textbf{34.4} & 47.7 & 27.3 & \textbf{87.3}
      & 74.1 & 63.3 & 36.5 & 96.6 & 44.7 \\
      & extended       & 34.0 & \textbf{48.3} & \textbf{27.7} & 83.3
      & \textbf{79.0} & \textbf{65.4} & \textbf{37.7} & 96.3 & \textbf{45.2} \\
      \bottomrule
  \end{tabular}
  \caption{\textbf{Prompt component sensitivity ablation} for
    Qwen3-VL-8B on OF-ItW. One-at-a-time variation over role persona,
    task instruction, label formatting, and class definitions with
    output format fixed to free text. Baseline variants marked
  $\ast$. Best per section \textbf{bold}.}
  \label{tab:component-ablation}
\end{table}
\endgroup

\paragraph{Model Scale Ablation.}
Table~\ref{tab:zeroshot-scale} reports zero-shot results on OF-ItW
for model sizes from 2B to 8B parameters across three MLLM families.
Performance generally increases with scale but plateaus around 8B
parameters. Models at 2B and 4B largely fail at \emph{fallen}
detection (F1 below 11), with InternVL3.5-2B as a notable exception
(Fallen F1 40.2). At the 8B level, all three models achieve
comparable fall detection F1 (67--68), while \emph{fallen} detection
remains substantially more challenging (F1 29--39).

\subsection{Trained Classifier Setup and Additional Results}\label{sec:app-implementation}

\begin{table}[t]
  \centering
  \renewcommand{\arraystretch}{0.95}
  \scriptsize
  \caption{In-distribution classification on OF-Staged (CV test splits).
    \textbf{B}al.\ \textbf{Acc}uracy, \textbf{Se}nsitivity,
  \textbf{Sp}ecificity.}
  \label{tab:unified-classification-cv}
  \begin{threeparttable}
    \setlength{\tabcolsep}{1.5pt}
    \begin{tabular}{l L L | CCC C CCC C CCC C CCE}
      \toprule
      \multirow{2}{*}{Test} &
      \multirow{2}{*}{Model} &
      \multirow{2}{*}{Train} &
      \multicolumn{3}{c}{16-class} & &
      \multicolumn{3}{c}{Fall $\Delta$} & &
      \multicolumn{3}{c}{Fallen $\Delta$} & &
      \multicolumn{3}{c}{Fall $\cup$ Fallen $\Delta$} \\
      \cmidrule(lr){4-6}\cmidrule(lr){8-10}\cmidrule(lr){12-14}\cmidrule(lr){16-18}
      & & & BAcc & Acc & F1 & &
      Se & Sp & F1 & &
      Se & Sp & F1 & &
      Se & Sp & F1 \\
      \midrule
      \multirow{2}{*}{\rotatebox{90}{\tiny{CMDF}}}
      & VMAE   & Sta     & 71.6 & 80.9 & \bst{73.9} && 94.1 & 95.7 & \bst{85.2} && 91.4 & 98.7 & \bst{90.8} && 96.0 & 94.6 & \bst{90.6} \\
      \oodrow
      & VMAE   & Syn     & 53.7 & 57.9 & \sbst{41.0} && 68.5 & 97.3 & \sbst{73.9} && 38.5 & 96.2 & \sbst{45.8} && 62.0 & 94.8 & \sbst{69.9} \\
      \addlinespace[1.6em]
      \midrule
      \addlinespace[0.4em]
      \oodrow
      \multirow{3}{*}{\rotatebox{90}{\tiny{UP-Fall}}}
      & VMAE   & CMDF    & 60.8 & 57.2 & \sbst{43.8} && 4.3 & 99.4 & 8.1 && 64.0 & 95.5 & \sbst{69.9} && 33.1 & 93.4 & \sbst{46.2} \\
      & VMAE   & Sta     & 56.5 & 89.1 & \bst{55.1} && 97.2 & 95.8 & \bst{91.3} && 99.1 & 93.2 & \bst{86.8} && 99.4 & 86.2 & \bst{90.2} \\
      \oodrow
      & VMAE   & Syn     & 49.6 & 56.3 & 29.9 && 43.1 & 99.9 & \sbst{60.1} && 5.3 & 99.4 & 9.8 && 26.0 & 99.6 & 41.1 \\
      \addlinespace[1em]
      \midrule
      \addlinespace[0.4em]
      \oodrow
      \multirow{3}{*}{\rotatebox{90}{\tiny{Le2i}}}
      & VMAE   & CMDF    & 59.1 & 72.3 & \sbst{50.2} && 86.5 & 95.5 & \sbst{80.0} && 96.8 & 94.4 & \sbst{80.0} && 92.6 & 88.8 & \sbst{81.3} \\
      & VMAE   & Sta     & 63.1 & 79.4 & \bst{64.8} && 97.3 & 99.2 & \bst{96.0} && 100 & 96.4 & \bst{87.3} && 100 & 95.3 & \bst{93.2} \\
      \oodrow
      & VMAE   & Syn     & 40.6 & 52.8 & 30.1 && 48.6 & 100 & 65.5 && 0.0 & 99.2 & 0.0 && 29.4 & 100 & 45.5 \\
      \addlinespace[0.4em]
      \midrule
      \addlinespace[0.4em]
      \oodrow
      \multirow{3}{*}{\rotatebox{90}{\tiny{GMDCSA}}}
      & VMAE   & CMDF    & 72.0 & 60.2 & \sbst{59.3} && 64.7 & 85.5 & 56.4 && 52.9 & 100 & \sbst{69.2} && 67.6 & 86.4 & \sbst{70.8} \\
      & VMAE   & Sta     & 79.4 & 79.6 & \bst{76.2} && 94.1 & 94.7 & \bst{86.5} && 70.6 & 98.7 & \bst{80.0} && 85.3 & 93.2 & \bst{86.6} \\
      \oodrow
      & VMAE   & Syn     & 73.4 & 61.3 & 48.4 && 76.5 & 97.4 & \sbst{81.2} && 11.8 & 100 & 21.1 && 47.1 & 98.3 & 62.7 \\
      \addlinespace[2em]
      \midrule
      \addlinespace[0.4em]
      \oodrow
      \multirow{3}{*}{\rotatebox{90}{\tiny{EDF}}}
      & VMAE   & CMDF    & 31.2 & 44.9 & \sbst{30.2} && 24.4 & 96.2 & \sbst{34.4} && 40.0 & 96.9 & \sbst{49.0} && 33.3 & 92.7 & 44.2 \\
      & VMAE   & Sta     & 45.8 & 63.0 & \bst{43.6} && 67.4 & 93.8 & \bst{68.2} && 83.3 & 94.6 & \bst{74.6} && 87.7 & 91.7 & \bst{84.2} \\
      \oodrow
      & VMAE   & Syn     & 31.0 & 41.3 & 22.7 && 32.6 & 94.3 & 40.6 && 60.0 & 87.9 & 48.0 && 64.4 & 86.7 & \sbst{65.3} \\
      \addlinespace[0.4em]
      \midrule
      \addlinespace[0.4em]
      \oodrow
      \multirow{3}{*}{\rotatebox{90}{\tiny{OCCU}}}
      & VMAE   & CMDF    & 42.0 & 41.4 & \sbst{29.9} && 14.3 & 99.5 & 24.5 && 40.0 & 96.0 & \sbst{50.0} && 26.8 & 94.3 & \sbst{38.9} \\
      & VMAE   & Sta     & 59.4 & 73.6 & \bst{50.1} && 88.1 & 94.4 & \bst{82.2} && 95.0 & 97.0 & \bst{90.5} && 92.7 & 89.8 & \bst{87.4} \\
      \oodrow
      & VMAE   & Syn     & 51.2 & 52.7 & 31.0 && 28.6 & 100 & \sbst{44.4} && 0.0 & 100 & 0.0 && 14.6 & 100 & 25.5 \\
      \addlinespace[0.8em]
      \midrule
      \addlinespace[0.4em]
      \oodrow
      \multirow{3}{*}{\rotatebox{90}{\tiny{MCFD}}}
      & VMAE   & CMDF    & 41.7 & 54.9 & \sbst{38.5} && 40.6 & 92.4 & \sbst{47.2} && 83.1 & 84.6 & \sbst{59.9} && 67.0 & 75.3 & \sbst{61.9} \\
      & VMAE   & Sta     & 47.3 & 67.4 & \bst{48.9} && 77.2 & 92.7 & \bst{74.1} && 70.0 & 94.9 & \bst{69.3} && 81.7 & 88.3 & \bst{79.4} \\
      \oodrow
      & VMAE   & Syn     & 31.1 & 40.5 & 22.9 && 41.5 & 93.1 & 48.5 && 7.1 & 98.5 & 12.3 && 29.4 & 90.9 & 39.8 \\
      \addlinespace[0.8em]
      \bottomrule
    \end{tabular}
    \begin{tablenotes}
      \scriptsize
    \item[\ ] VMAE: VideoMAE-K400~\cite{tong2022videomae}.
    \item[\ ] CMDF: CMDFall~CS,
      Sta: OF-Sta~CV,
      Syn: OF-Syn~Ra.
    \item[\ ] \colorbox{blue!6}{Shaded} rows: out-of-distribution (test data absent from training).
    \end{tablenotes}
  \end{threeparttable}
\end{table}

\paragraph{Feature Extraction.}\label{sec:app-features}
We extract features from three backbone networks to enable systematic
evaluation of visual representations for fall detection. The SigLIP2 features were also used for the timeline segmentation experiments.
Table~\ref{tab:feature-settings} summarizes the extraction
configuration for each backbone.

\begin{table}[ht]
  \centering
  \setlength{\tabcolsep}{2pt}
  \small
  \caption{Feature extraction configurations. This table details the
    precise settings used for each backbone network to ensure
  reproducibility of our results, as referenced in Section~\ref{sec:exp}.}
  \label{tab:feature-settings}
  \renewcommand{\arraystretch}{1.2}
  \begin{tabular}{lccccccc}
    \toprule
    \textbf{Backbone} & \textbf{Stride} & \textbf{FPS} &
    \textbf{Frames} & \textbf{Span (s)} & \textbf{Dim.} & \textbf{Pad} \\
    \midrule
    I3D~\cite{carreira2017quo} & $\frac{1}{\text{fps}}$ & original
    & 16 & $16/\text{fps}$ & 2048 & Yes \\
    SigLip2~\cite{tschannen2025siglip} & $\frac{1}{\text{fps}}$ &
    original & 1 & $1/\text{fps}$ & 768 & - \\
    VMAE~\cite{tong2022videomae} & $\frac{2}{15}$ & 7.5 fps & 16 &
    2.13 & 768 & No \\
    \bottomrule
  \end{tabular}
\end{table}

Features are extracted following model-specific protocols:
\begin{itemize}
  \item \textbf{I3D:} Extracted with frame-wise stride and 16-frame
    windows. Features are centered on each frame with padding at
    video boundaries.

  \item \textbf{SigLIP2:} For each video frame at its original frame
    rate, we extract a 768-D SigLIP2 representation using the
    standard encoder without temporal context.

  \item \textbf{VideoMAE} Extracted at 7.5 fps with
    $\frac{2}{15}$-second stride, using 16-frame windows. Features
    are obtained by averaging the token dimension of the final hidden state.
\end{itemize}


\paragraph{Computational Resources.}
Our experiments were conducted on four NVIDIA H100 GPUs, though they
could be run on a single GPU.



\paragraph{Unsupervised baseline results for timeline segmentation.}

We report results based on clustering-based unsupervised timeline
segmentation in~\Cref{tab:segmentation-core-unsup}.

\begin{table}[t]
  \centering
  \caption{Unsupervised action segmentation on three representative domains.}
  \label{tab:segmentation-core-unsup}
  \setlength{\tabcolsep}{2pt}
  \scriptsize
  \begin{tabular}{l l cccc c}
    \toprule
    \textbf{Domain} & \textbf{Method} & \textbf{F1@10} & \textbf{F1@25} & \textbf{F1@50} & \textbf{Edit} & \textbf{Acc} \\
    \midrule

    \multirow{3}{*}{\textbf{CMDFall (staged)}}

    & TW\mbox{-}FINCH & \bst{64.8} & \bst{60.0} & \bst{43.4} & \bst{59.7} & \bst{64.8} \\
    & KMeans          & 48.4 & 44.0 & 29.8 & 42.1 & 70.0 \\

    \addlinespace[0.8ex]
    \midrule

    \multirow{3}{*}{\textbf{OF\mbox{-}ItW (in-the-wild)}}
    & TW\mbox{-}FINCH & \bst{67.1} & \bst{63.4} & \bst{42.1} & \bst{54.8} & \bst{59.8} \\
    & KMeans          & 58.5 & 55.1 & 35.9 & 50.6 & 57.9 \\

    \addlinespace[0.8ex]
    \midrule

    \multirow{3}{*}{\textbf{OF\mbox{-}Syn (synthetic)}}
    & TW\mbox{-}FINCH & \bst{81.7} & \bst{80.8} & \bst{71.5} & \bst{71.2} & \bst{76.6} \\
    & KMeans          & 77.0 & 78.3 & 67.6 & 69.3 & 78.4 \\

    \addlinespace[0.8ex]
    \bottomrule
  \end{tabular}
\end{table}

\subsubsection{Cross-Demographic Generalization}

\begin{figure}[t]
  \centering
  \includegraphics[width=0.8\linewidth]{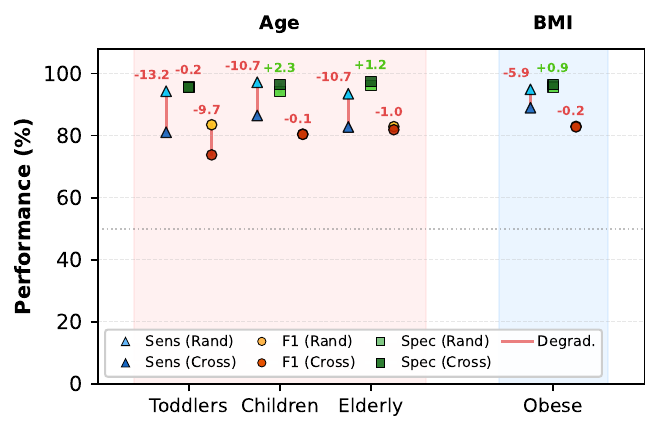}
  \caption{OF-Syn demographic splits; \emph{sensitivity},
  \emph{specificity}, \emph{F1}}
  \label{fig:subgroup-comparison}
\end{figure}

\paragraph{Cross-attribute splits.}
We evaluate cross-age and cross-BMI training on
OF-Syn in Figure~\ref{fig:subgroup-comparison} and report metrics for
the class \emph{Fall}. We train on the age groups \emph{young
adult} and \emph{middle aged} and evaluate on \emph{toddlers},
\emph{children}, and \emph{elderly}. A marked drop in
sensitivity is observed, and for toddlers, also in F1 score. Slight negative
effects are also visible for cross-BMI training.

\subsubsection{Analysis of Domain Shift Results}

\paragraph{Numerical Values for Fr\'{e}chet Video Distances.}\label{sec:app-fvd}
Figure~\ref{fig:domain-heatmaps} in Section~\ref{sec:eval-domain}
visualizes pairwise Fr\'{e}chet Video Distances (FVD) between datasets.
Tables~\ref{tab:fvd-i3d}--\ref{tab:fvd-siglip2} provide the exact
numerical values for all three backbone networks. 


\begin{table}[ht]
  \centering
  \scriptsize
  \caption{Fr\'{e}chet Video Distances between datasets using \textbf{I3D
  features}. Lower values indicate higher similarity between domains.}
  \label{tab:fvd-i3d}
  \setlength{\tabcolsep}{2pt} 
  \renewcommand{\arraystretch}{0.95} 
  \scalebox{0.8}{
    \begin{tabular}{l|c|c|c|c|c|c|c|c|c|c}
      \toprule
      & CMDF & UP Fall & Le2i & GMDCSA & EDF & OCCU & Cauca & MCFD &
      OF-ItW & OF-Syn \\
      \midrule
      CMDF & 0.00 & 88.60 & 77.06 & 73.86 & 100.86 & 104.61 & 91.86 &
      101.88 & 103.60 & 99.37\\
      UP Fall & 88.60 & 0.00 & 114.54 & 101.89 & 107.14 & 116.99 &
      118.38 & 115.95 & 114.13 & 108.55\\
      Le2i & 77.06 & 114.54 & 0.00 & 104.66 & 119.73 & 134.99 &
      124.16 & 126.10 & 130.03 & 120.50\\
      GMDCSA & 73.86 & 101.89 & 104.66 & 0.00 & 98.52 & 110.26 &
      97.87 & 123.29 & 110.01 & 100.45\\
      EDF & 100.86 & 107.14 & 119.73 & 98.52 & 0.00 & 102.20 & 117.09
      & 116.89 & 133.00 & 117.07\\
      OCCU & 104.61 & 116.99 & 134.99 & 110.26 & 102.20 & 0.00 &
      117.92 & 134.54 & 163.98 & 156.36\\
      Cauca & 91.86 & 118.38 & 124.16 & 97.87 & 117.09 & 117.92 &
      0.00 & 128.56 & 134.11 & 134.49\\
      MCFD & 101.88 & 115.95 & 126.10 & 123.29 & 116.89 & 134.54 &
      128.56 & 0.00 & 140.65 & 129.39\\
      OF-ItW & 103.60 & 114.13 & 130.03 & 110.01 & 133.00 & 163.98 &
      134.11 & 140.65 & 0.00 & 61.12\\
      OF-Syn & 99.37 & 108.55 & 120.50 & 100.45 & 117.07 & 156.36 &
      134.49 & 129.39 & 61.12 & 0.00 \\
      \bottomrule
  \end{tabular}}
\end{table}

\begin{table}[ht]
  \centering
  \scriptsize
  \caption{Fr\'{e}chet Video Distances between datasets using
    \textbf{VideoMAE features}. Lower values indicate higher
  similarities between domains.}
  \label{tab:fvd-vmae-pre}
  \setlength{\tabcolsep}{2pt} 
  \renewcommand{\arraystretch}{0.95} 
  \scalebox{0.8}{
    \begin{tabular}{l|c|c|c|c|c|c|c|c|c|c}
      \toprule
      & CMDF & UP Fall & Le2i & GMDCSA & EDF & OCCU & Cauca & MCFD &
      OF-ItW & OF-Syn \\
      \midrule
      CMDF   & 0.0 & 3521.8 & 2365.0 & 3042.0 & 3146.9 & 4178.9 &
      3253.5 & 3258.0 & 4060.6 & 2941.6 \\
      UP Fall& 3521.8 & 0.0 & 3407.9 & 3492.3 & 3325.7 & 4115.7 &
      4049.1 & 3276.3 & 4980.1 & 3430.0 \\
      Le2i   & 2365.0 & 3407.9 & 0.0 & 3438.2 & 2978.1 & 3763.5 &
      3967.0 & 3016.3 & 4842.5 & 3789.6 \\
      GMDCSA & 3042.0 & 3492.3 & 3438.2 & 0.0 & 3520.3 & 3798.4 &
      3380.6 & 3757.2 & 4466.7 & 3326.2 \\
      EDF    & 3146.9 & 3325.7 & 2978.1 & 3520.3 & 0.0 & 2991.4 &
      3918.6 & 3133.9 & 5024.1 & 3679.0 \\
      OCCU   & 4178.9 & 4115.7 & 3763.5 & 3798.4 & 2991.4 & 0.0 &
      4535.3 & 3366.0 & 5701.8 & 4976.3 \\
      Cauca  & 3253.5 & 4049.1 & 3967.0 & 3380.6 & 3918.6 & 4535.3 &
      0.0 & 3885.5 & 5204.8 & 4137.8 \\
      MCFD   & 3258.0 & 3276.3 & 3016.3 & 3757.2 & 3133.9 & 3366.0 &
      3885.5 & 0.0 & 4554.3 & 3955.5 \\
      OF-ItW & 4060.6 & 4980.1 & 4842.5 & 4466.7 & 5024.1 & 5701.8 &
      5204.8 & 4554.3 & 0.0 & 2289.1 \\
      OF-Syn & 2941.6 & 3430.0 & 3789.6 & 3326.2 & 3679.0 & 4976.3 &
      4137.8 & 3955.5 & 2289.1 & 0.0 \\
      \bottomrule
  \end{tabular}}
\end{table}

\begin{table}[ht]
  \centering
  \scriptsize
  \caption{Fr\'{e}chet Video Distances between datasets using
  \textbf{SigLIP2 features}.}
  \label{tab:fvd-siglip2}
  \setlength{\tabcolsep}{2pt} 
  \renewcommand{\arraystretch}{0.95} 
  \scalebox{0.85}{
    \begin{tabular}{l|c|c|c|c|c|c|c|c|c|c}
      \toprule
      & CMDF & UP Fall & Le2i & GMDCSA & EDF & OCCU & Cauca & MCFD &
      OF-ItW & OF-Syn \\
      \midrule
      CMDF & 0.00 & 0.40 & 0.39 & 0.43 & 0.41 & 0.45 & 0.46 & 0.40 &
      0.66 & 0.62\\
      UP Fall & 0.40 & 0.00 & 0.49 & 0.63 & 0.50 & 0.53 & 0.61 & 0.44
      & 0.74 & 0.65\\
      Le2i & 0.39 & 0.49 & 0.00 & 0.48 & 0.47 & 0.49 & 0.42 & 0.42 &
      0.65 & 0.69\\
      GMDCSA & 0.43 & 0.63 & 0.48 & 0.00 & 0.60 & 0.59 & 0.45 & 0.58
      & 0.66 & 0.70\\
      EDF & 0.41 & 0.50 & 0.47 & 0.60 & 0.00 & 0.20 & 0.53 & 0.33 &
      0.70 & 0.73\\
      OCCU & 0.45 & 0.53 & 0.49 & 0.59 & 0.20 & 0.00 & 0.50 & 0.35 &
      0.73 & 0.77\\
      Cauca & 0.46 & 0.61 & 0.42 & 0.45 & 0.53 & 0.50 & 0.00 & 0.37 &
      0.54 & 0.71\\
      MCFD & 0.40 & 0.44 & 0.42 & 0.58 & 0.33 & 0.35 & 0.37 & 0.00 &
      0.62 & 0.69\\
      OF-ItW & 0.66 & 0.74 & 0.65 & 0.66 & 0.70 & 0.73 & 0.54 & 0.62 &
      0.00 & 0.40\\
      OF-Syn & 0.62 & 0.65 & 0.69 & 0.70 & 0.73 & 0.77 & 0.71 & 0.69 &
      0.40 & 0.00 \\
      \bottomrule
    \end{tabular}
  }
\end{table}

\subsubsection{Feature-space visualization}\label{sec:app-feature-viz}
The feature-space projections shown in the main paper
(Figure~\ref{fig:tsne-hnne}) use
t\mbox{-}SNE~\cite{van2008visualizing} and
h\mbox{-}NNE~\cite{sarfraz2022hierarchical}. h\mbox{-}NNE provides
both a 2D embedding and a multi\mbox{-}scale partitioning, stabilizing
cluster contours and exposing local\,$\rightarrow$\,global
organization. The plots jointly make the cross\mbox{-}domain gap
explicit and indicate how much action structure is already present in
frozen features.
Below, we provide individual full-size visualizations of these
embeddings across backbones and dimensionality reduction techniques.

\newcommand{\imgscale}{0.75}

\begin{figure*}
  \centering
  \includegraphics[width=\imgscale\linewidth]{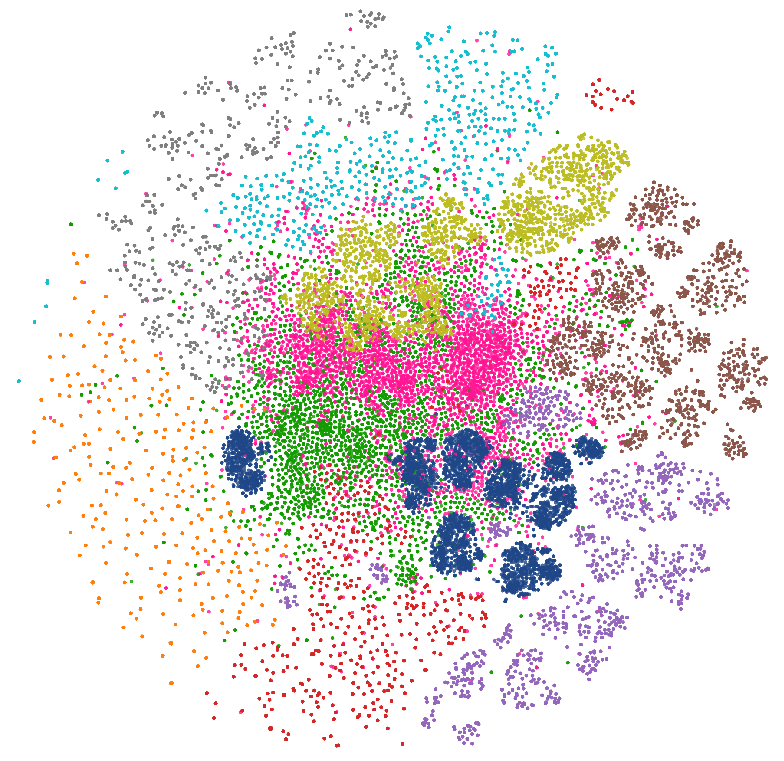}
  \includegraphics[width=\imgscale\linewidth,trim=0 0 320
  0,clip]{imgs/labels2.pdf}

  \caption{I3D-extracted 2D projections using TSNE, colour-coded by dataset.}
  \label{fig:appendix_ID3_TSNE_Data}
\end{figure*}
\begin{figure*}
  \centering
  \includegraphics[width=\imgscale\linewidth]{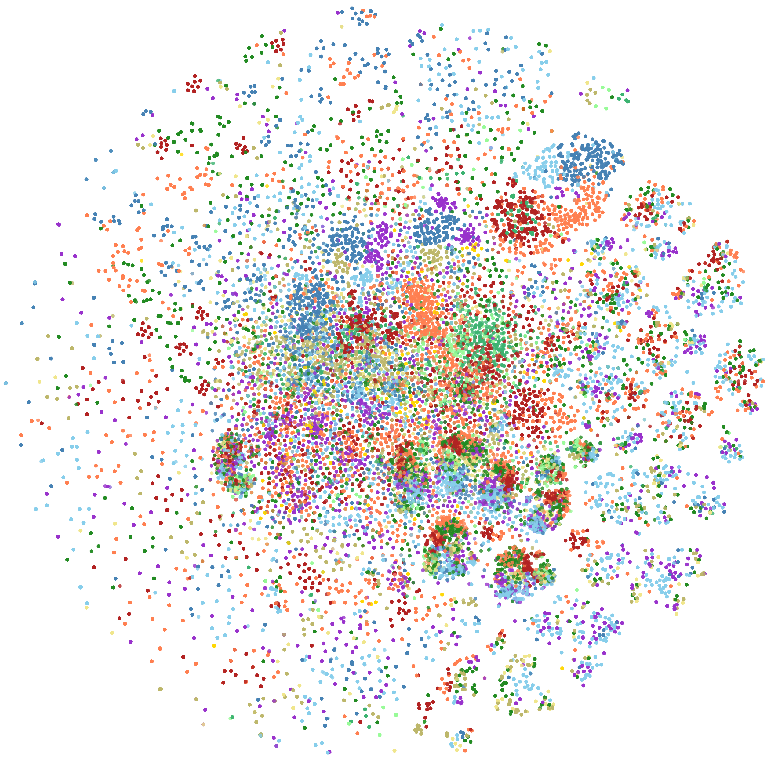}
  \includegraphics[width=\imgscale\linewidth,trim=300 0 0
  0,clip]{imgs/labels2.pdf}

  \caption{I3D-extracted 2D projections using TSNE, colour-coded by activity.}
  \label{fig:appendix_ID3_TSNE_Activity}
\end{figure*}

\begin{figure*}
  \centering
  \includegraphics[width=\imgscale\linewidth]{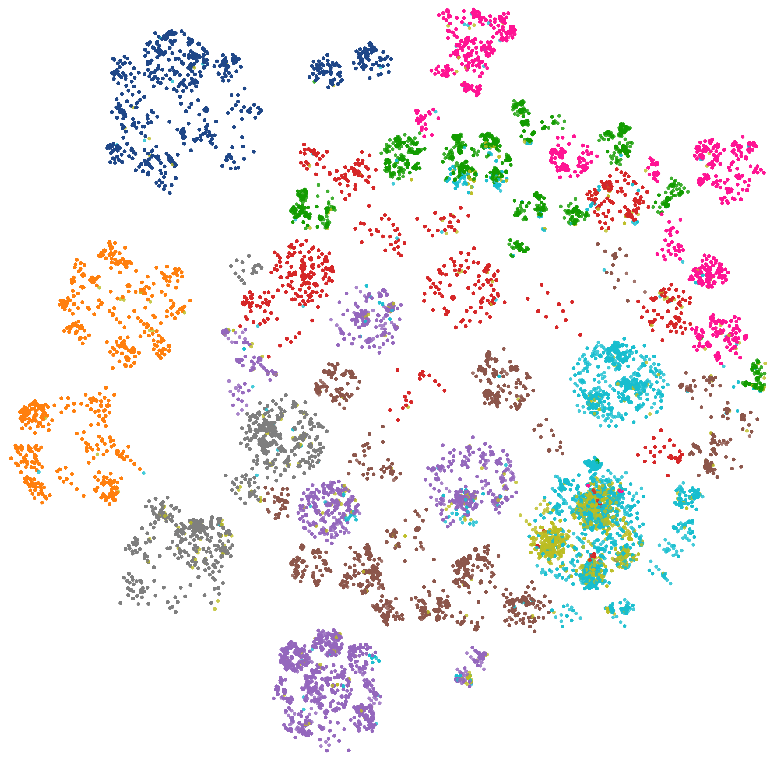}
  \includegraphics[width=\imgscale\linewidth,trim=0 0 320
  0,clip]{imgs/labels2.pdf}

  \caption{I3D-extracted 2D projections using HNNE, colour-coded by dataset.}
  \label{fig:appendix_ID3_HNNE_Data}
\end{figure*}
\begin{figure*}
  \centering
  \includegraphics[width=\imgscale\linewidth]{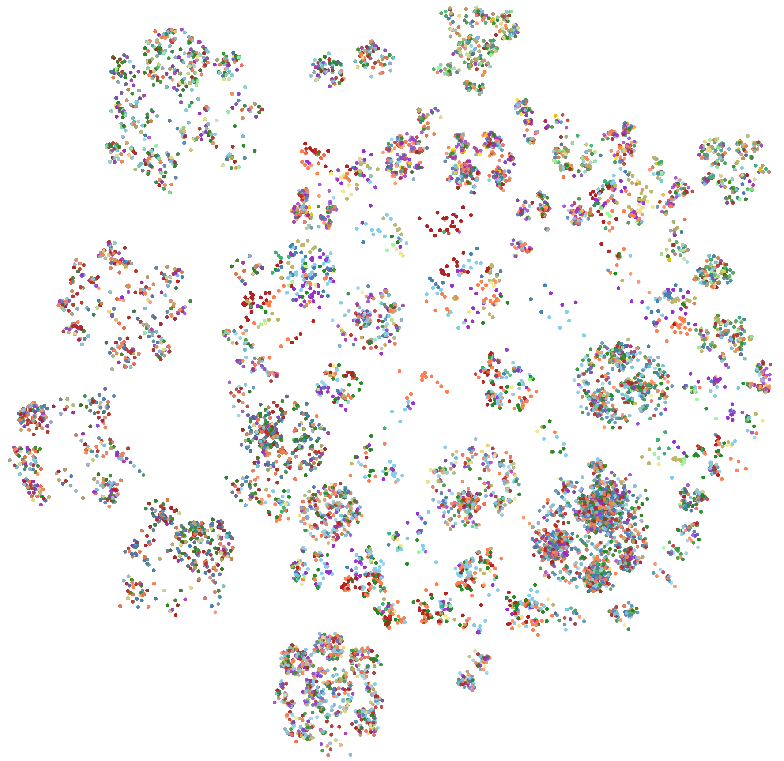}
  \includegraphics[width=\imgscale\linewidth,trim=300 0 0
  0,clip]{imgs/labels2.pdf}

  \caption{I3D-extracted 2D projections using HNNE, colour-coded by activity.}
  \label{fig:appendix_ID3_HNNE_Activity}
\end{figure*}


\begin{figure*}
  \centering
  \includegraphics[width=\imgscale\linewidth]{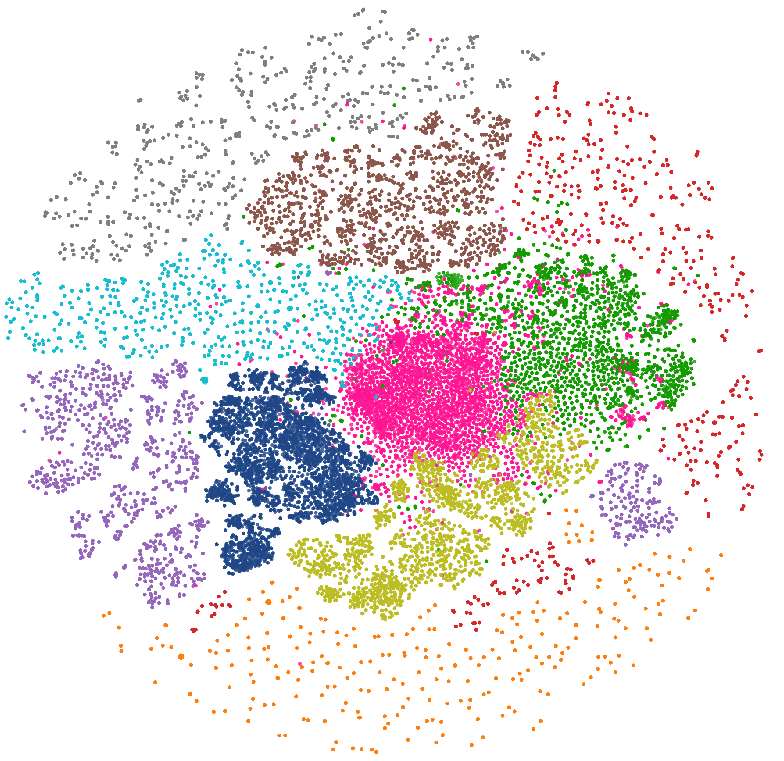}
  \includegraphics[width=\imgscale\linewidth,trim=0 0 320
  0,clip]{imgs/labels2.pdf}

  \caption{VMAE-extracted 2D projections using TSNE, colour-coded by dataset.}
  \label{fig:appendix_VMAE_TSNE_Data}
\end{figure*}
\begin{figure*}
  \centering
  \includegraphics[width=\imgscale\linewidth]{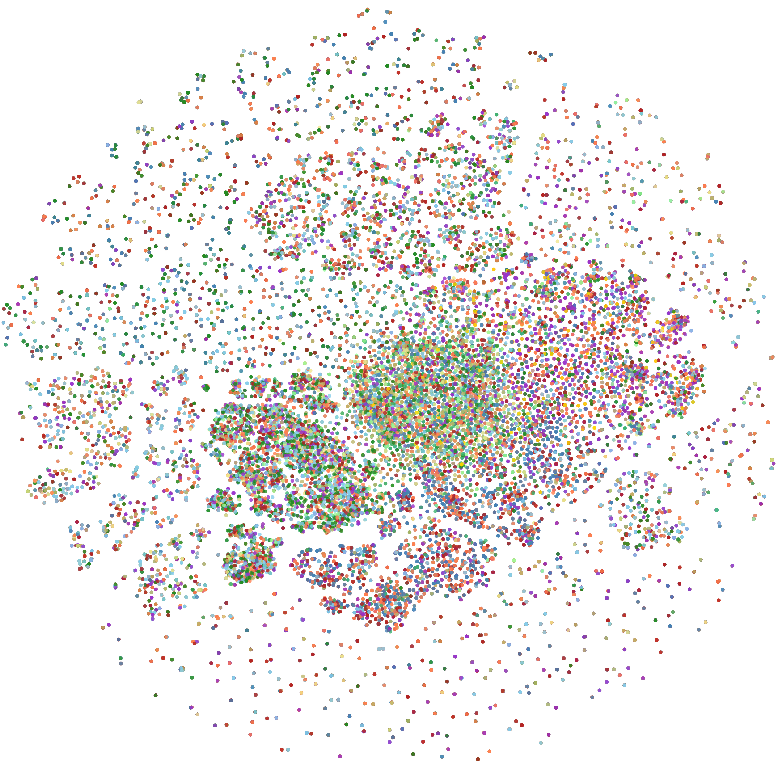}
  \includegraphics[width=\imgscale\linewidth,trim=300 0 0
  0,clip]{imgs/labels2.pdf}

  \caption{VMAE-extracted 2D projections using TSNE, colour-coded by activity.}
  \label{fig:appendix_VMAE_TSNE_Activity}
\end{figure*}

\begin{figure*}
  \centering
  \includegraphics[width=\imgscale\linewidth]{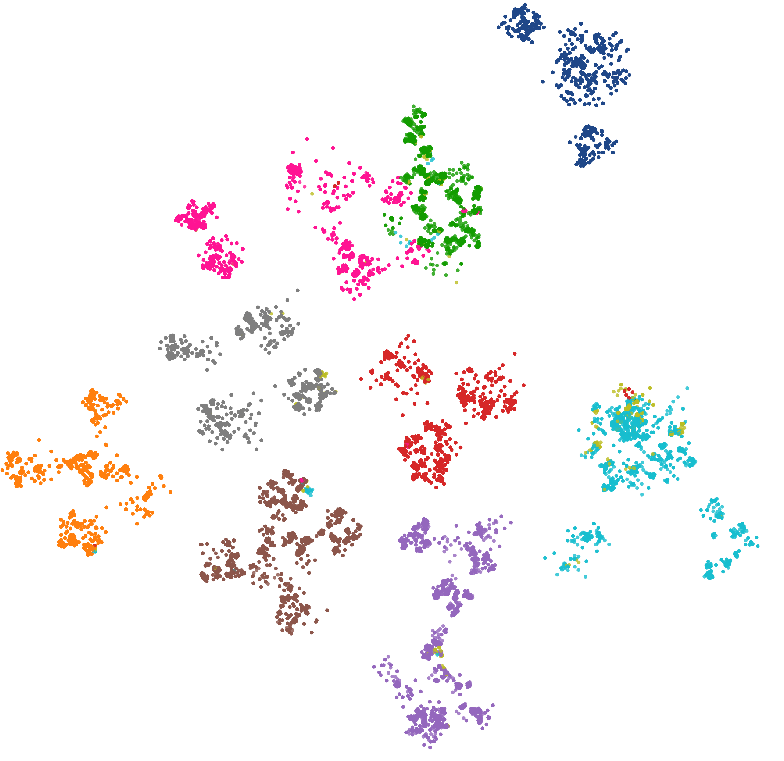}
  \includegraphics[width=\imgscale\linewidth,trim=0 0 320
  0,clip]{imgs/labels2.pdf}

  \caption{VMAE-extracted 2D projections using HNNE, colour-coded by dataset.}
  \label{fig:appendix_VMAE_HNNE_Data}
\end{figure*}
\begin{figure*}
  \centering
  \includegraphics[width=\imgscale\linewidth]{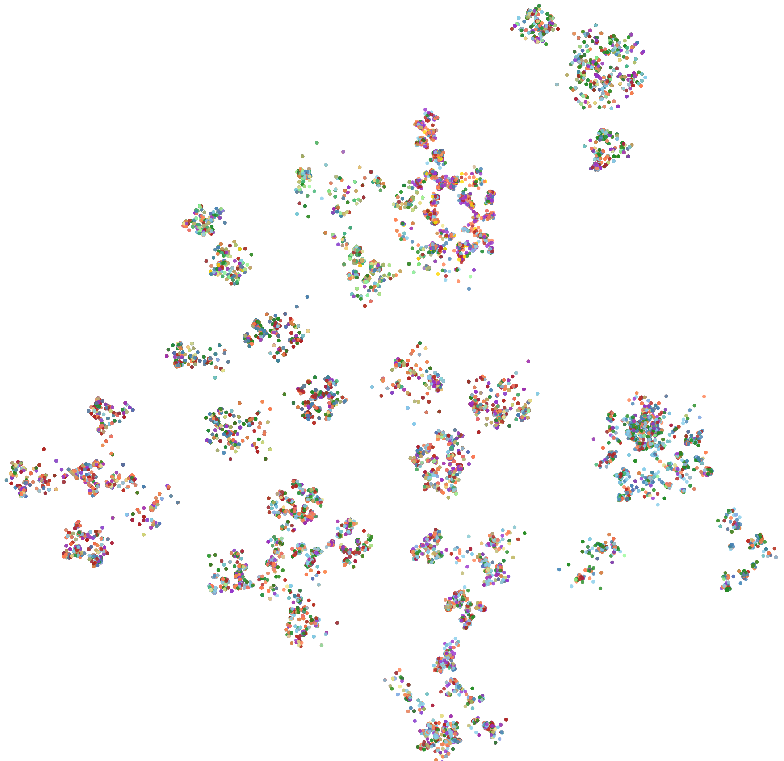}
  \includegraphics[width=\imgscale\linewidth,trim=300 0 0
  0,clip]{imgs/labels2.pdf}

  \caption{VMAE-extracted 2D projections using HNNE, colour-coded by activity.}
  \label{fig:appendix_VMAE_HNNE_Activity}
\end{figure*}


\begin{figure*}
  \centering
  \includegraphics[width=\imgscale\linewidth]{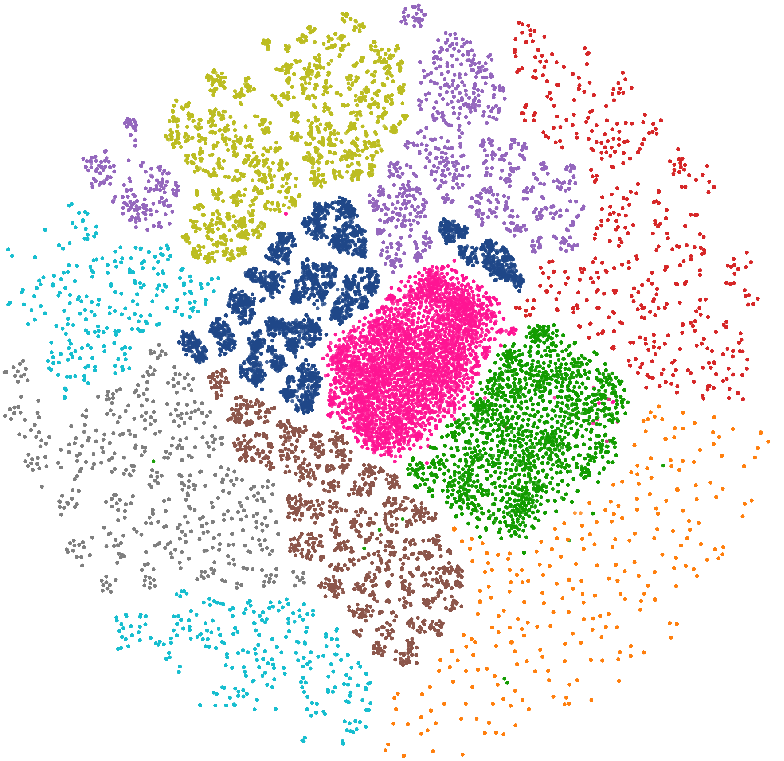}
  \includegraphics[width=\imgscale\linewidth,trim=0 0 320
  0,clip]{imgs/labels2.pdf}

  \caption{SigLIP2-extracted 2D projections using TSNE, colour-coded
  by dataset.}
  \label{fig:appendix_SigLIP2_TSNE_Data}
\end{figure*}
\begin{figure*}
  \centering
  \includegraphics[width=\imgscale\linewidth]{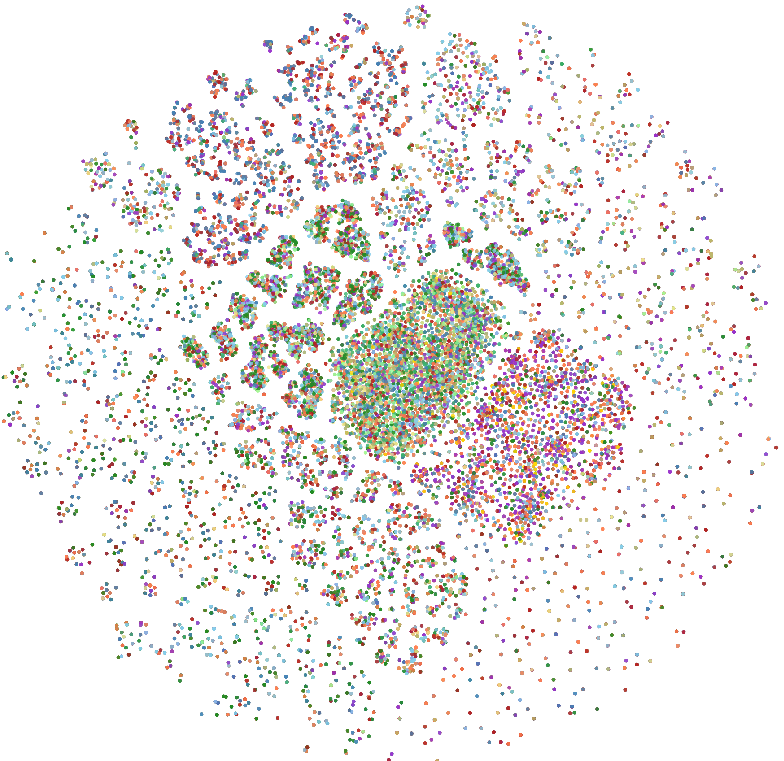}
  \includegraphics[width=\imgscale\linewidth,trim=300 0 0
  0,clip]{imgs/labels2.pdf}

  \caption{SigLIP2-extracted 2D projections using TSNE, colour-coded
  by activity.}
  \label{fig:appendix_SigLIP2_TSNE_Activity}
\end{figure*}

\begin{figure*}
  \centering
  \includegraphics[width=\imgscale\linewidth]{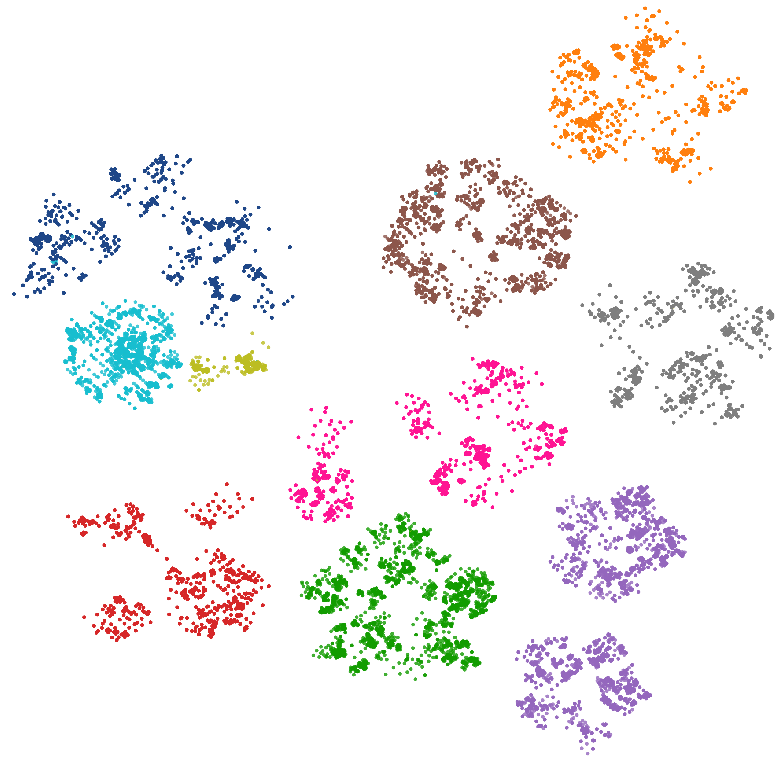}
  \includegraphics[width=\imgscale\linewidth,trim=0 0 320
  0,clip]{imgs/labels2.pdf}

  \caption{SigLIP2-extracted 2D projections using HNNE, colour-coded
  by dataset.}
  \label{fig:appendix_SigLIP2_HNNE_Data}
\end{figure*}
\begin{figure*}
  \centering
  \includegraphics[width=\imgscale\linewidth]{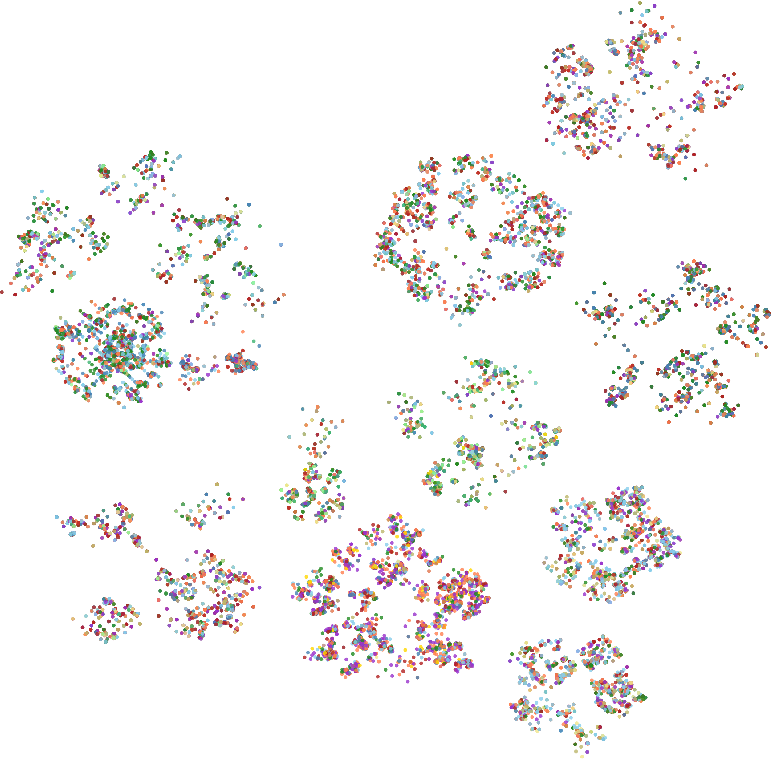}
  \includegraphics[width=\imgscale\linewidth,trim=300 0 0
  0,clip]{imgs/labels2.pdf}

  \caption{SigLIP2-extracted 2D projections using HNNE, colour-coded
  by activity.}
  \label{fig:appendix_SigLIP2_HNNE_Activity}
\end{figure*}


\end{document}